\documentclass[letterpaper, 10 pt, conference]{ieeeconf}  

\IEEEoverridecommandlockouts                              

\usepackage{graphics}
\usepackage{amsfonts}
\usepackage{amsmath}

\usepackage{amssymb}
\usepackage[colorlinks, linkcolor=red, citecolor=blue]{hyperref}
\usepackage{algorithm}
\usepackage{algpseudocode}
\usepackage{epsfig}
\usepackage{graphicx}
\usepackage{caption}
\usepackage{subfigure}
\usepackage{mathrsfs}
\usepackage{array}
\usepackage{multirow}
\usepackage{multicol}
\usepackage{booktabs}
\usepackage{makecell}
\usepackage{xcolor}
\usepackage{amsfonts}
\usepackage{gensymb}
\usepackage{flushend}
\usepackage{algorithm}
\usepackage{algorithmicx}
\usepackage{algpseudocode}
\usepackage{amsmath}

\title{\LARGE \bf
A Hyper-network Based End-to-end Visual Servoing \\
with Arbitrary Desired Poses
}

\author{Hongxiang Yu$^{1}$, Anzhe Chen$^{1}$, Kechun Xu$^{1}$, Zhongxiang Zhou$^{1}$, Wei Jing$^{2}$, Yue Wang$^{1}$ and Rong Xiong$^{1}$
}

\begin{document}

\maketitle
\thispagestyle{empty}
\pagestyle{empty}

\begin{abstract}

Recently, several works achieve end-to-end visual servoing (VS) for robotic manipulation by replacing traditional controller with differentiable neural networks, but lose the ability to servo arbitrary desired poses. This letter proposes a differentiable architecture for arbitrary pose servoing: a hyper-network based neural controller (HPN-NC). To achieve this, HPN-NC consists of a hyper net and a low-level controller, where the hyper net learns to generate the parameters of the low-level controller and the controller uses the 2D keypoints error for control like traditional image-based visual servoing (IBVS). HPN-NC can complete 6 degree of freedom visual servoing with large initial offset. Taking advantage of the fully differentiable nature of HPN-NC, we provide a three-stage training procedure to servo real world objects. With self-supervised end-to-end training, the performance of the integrated model can be further improved in unseen scenes and the amount of manual annotations can be significantly reduced.
 

\end{abstract}

\section{INTRODUCTION}

Visual servoing (VS) is a technique that uses vision feedback to guide the robot to achieve high-precision positioning. In classical VS\cite{chaumette2006basic,bakthavatchalam2013photometric,shi2018adaptive, wang2018adaptive}, a set of handcrafted visual features such as points, lines, contours, moments are extracted and compared with features of a pre-defined desired pose. A manually designed controller then moves the camera to the pre-defined desired pose by reducing the feature error between the desired and current poses.

With the development of deep learning, some learning-based methods have emerged to reduce the excessive manual effort in VS. \cite{saxena2017exploring,bateux2018icra,yu2019siamese,felton2021siame} use Convolutional Neural Networks (CNNs) to process the images observed at the current and the desired pose separately and estimate the relative pose, following by a position-based visual servo (PBVS) controller\cite{chaumette2006basic}. They get rid of expensive manual feature annotations by considering the whole image as a feature. However, the pose estimation performance depends on the similarity of the input image pairs, so the offset between the initial pose and the desired pose cannot be large \cite{adrian2022dfbvs}. \cite{adrian2022dfbvs,harish2020dfvs,katara2021deepmpcvs} use deep learning to improve the reliability of 2D correspondences extraction followed by an image-based visual servo (IBVS) controller\cite{chaumette2006basic} or IBVS-based MPC\cite{katara2021deepmpcvs}. The consistency of features enables VS with large pose offset. But IBVS has inherent drawbacks such as small convergence region and local minima\cite{espiau1992new,kelly2000stable}.

\begin{figure}[t]
\centering
\includegraphics[width=1\linewidth]{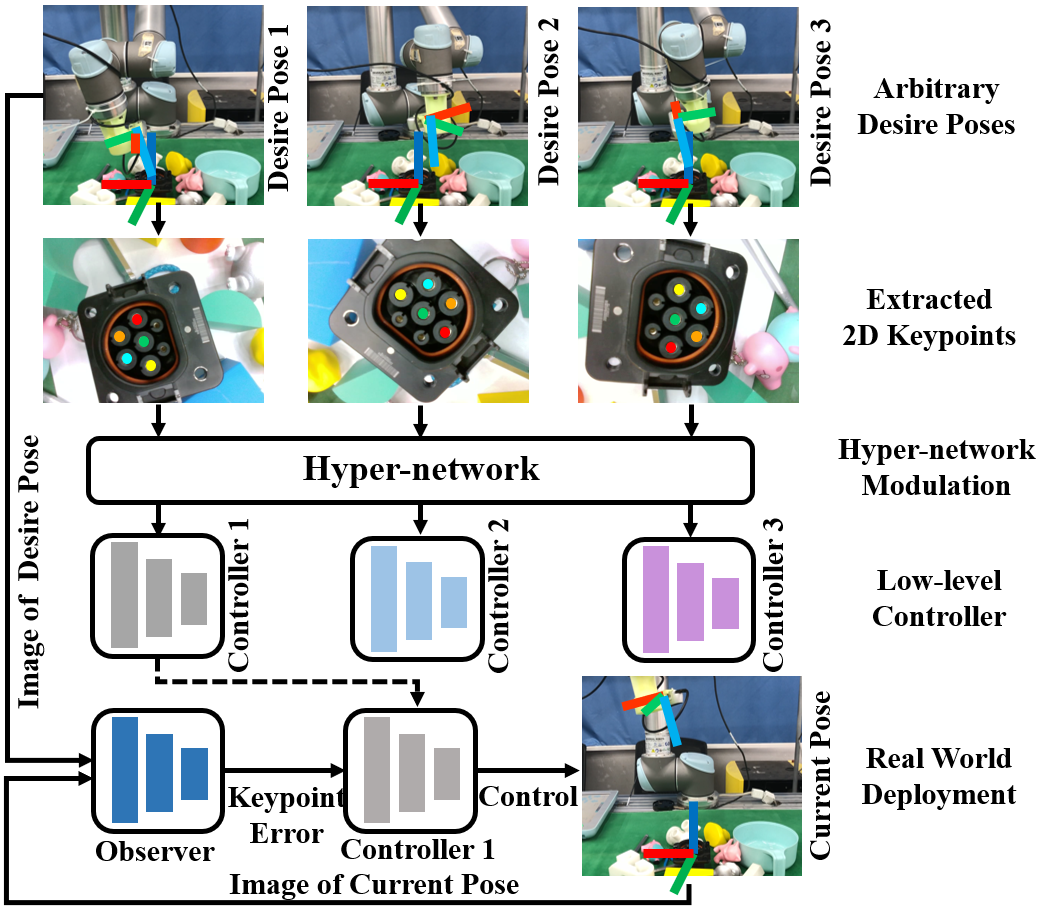}
\caption{The upper coordinate represents the camera's frame and the lower represents the target object's frame. Given the keypoints of desired poses, HPN generates an unique neural controller for each pose. For example, given the keypoints observed at desired Pose 1, HPN generates Controller 1 that helps the camera move to Pose 1 from arbitrary initial poses.}
\label{fig:fig1_teaser}
\vspace{-0.7cm}
\end{figure}

Recently, several works achieve self-supervised end-to-end VS for robotic manipulation tasks\cite{puang2020kovis,wang2022end,levine2016end}. Given a real world target object, they take the 2D keypoints extracted in an unsupervised manner as features, which avoids either manual keypoint annotations and pose estimation. By replacing IBVS with a neural controller, they make the whole architecture differentiable which enables self-supervised end-to-end training. 
However, these methods have an obvious weakness in that they are designed to servo a fixed desired pose , thus losing the functionality of the traditional VS controller to handle arbitrary desired poses. By only taking the image observed at the current pose as input, they lack the information about the desired pose. This means that if the number of desired poses is limited, they will have to train several neural controllers and select the appropriate one according to the given desired pose. But considering a scenario that needs to change the desired pose constantly, these methods will have to train infinite controllers or to frequently fine-tune the controller to ensure VS performance. \emph{How to implement a lightweight differentiable neural controller that can servo arbitrary desired poses remains a challenging problem.}

In this letter, we try to investigate an appropriate neural controller architecture capable of servoing arbitrary desired poses. To servo a random sampled desired poses in the 6 degree of freedom (DOF) space, it is not practical to train an infinite number of controllers. 
Simply adding an input that encoding information of the desired pose will inevitably enlarge network volume and prolong the inference time. In contrast, we define servoing a single desired pose as a task, and state servoing arbitrary desired poses as a multi-task learning problem. We proposing a hyper-network (HPN)\cite{ha2016hypernetworks} based neural controller (HPN-NC) to tackle this problem. HPN-NC consists of a hyper net and a low-level neural controller. Following \cite{puang2020kovis,wang2022end,levine2016end}, we use 2D keypoints as features. As shown in Fig.~\ref{fig:fig1_teaser}, given 2D keypoints extracted at a desired pose, the hyper net generates the parameters for the low-level controller corresponding to this desired pose. The modulated low-level controller takes the error between current and desired 2D keypoints for control inference. In this way, we can generate an unique controller for each desired pose that avoids endless fine-tuing. HPN-NC outperforms traditional IBVS and other neural controllers(NCs). To servo real world objects, we connect it with a supervised neural observer(NO) and provide a three-stage training procedure: training HPN-NC in simulation with synthetic data, training NO with manual keypoint annotations and end-to-end training the integrated model(IM:NO with NC) with all synthetic, annotated and robot's self-supervision data. Taking advantage of the fully differentiable nature of HPN-NC, IM can be further improved fully automatically in an end-to-end manner transfering to unseen scenes. Note that the amount of manual annotations can also be significantly reduced through end-to-end training, as they are only used as regularizer.
Overall, the contributions of this paper are three-fold:
\begin{itemize}
\item  We are the first work to state VS arbitrary desired poses with neural network as a multi-task learning problem. To solve this problem, we propose the HPN-NC. It outperforms other network structures when VS arbitrary desired poses both in simulation and real world experiments.

\item Neural controllers can be further fine-tuned fully automatically with self-supervised end-to-end training. HPN-NC's fine-tuning ability outperforms other NCs given error-free 2D keypoints in simulation. Given imprecise 2D keypoints in unseen scene in real world, the performance of IM consists of NO and HPN-NC can be further improved.

\item We also consider a situation with insufficient manual annotations. Self-supervised end-to-end training enables IM
to achieve 92$\%$ success rate VS with only 30 annotations.


\end{itemize}




\section{RELATED WORK}

\textbf{Traditional VS Methods:} 
Classical VS methods can help the robot achieve high-precision positioning through vision feedback but highly rely on handcrafted visual features, manually labeling or recognizable QR code \cite{shi2018adaptive, wang2018adaptive}
. Traditional VS controllers include IBVS\cite{chaumette2006basic,allibert2010predictive}, PBVS\cite{thuilot2002pbvs,park2012pbvs} and hybrid approaches \cite{gans2007switch,hafez2007weighted}. PBVS uses the relative pose between current and desired pose as visual feature and plans a globally asymptotically stable straight trajectory in 3D Cartesian space. IBVS uses matched keypoints on 2D image plane, which is insensitive to calibration error, but suffers from small convergence region due to the high non-linearity \cite{espiau1992new,kelly2000stable}. It may meet feature loss problem\cite{jin2021policy} dealing with large initial pose offset.

\textbf{Deep learning Based Methods:} Deep Neural Networks\cite{krizhevsky2012imagenet, huang2017densely} have shown remarkable feature extraction ability on various tasks such as detection, segmentation or tracking, and also alleviate the dependency of manual effort for VS task. 

Pose estimation methods\cite{bateux2018icra,saxena2017exploring, yu2019siamese} usually bypass the 2D keypoints prediction and estimate the relative pose or directly to predict control commands from image pairs observed at the current and the desired poses. \cite{bateux2018icra} implements a deep nerual network to estimate relative pose between current camera pose and the desired camera pose, then performs PBVS based on the relative pose. \cite{saxena2017exploring} trains a convolutional neural network over the whole image with synchronised camera poses to guide the quadrotor. \cite{yu2019siamese} proposes a new neural network based on a Siamese architecture which outputs the relative pose between any pair of images and realize VGA connector insertion with submillimeter accuracy. As the input image pairs have to be similar to ensure the performance of learning based pose estimator, camera pose offsets between the desired and the initial poses are limited.

Keypoint based methods\cite{adrian2022dfbvs,harish2020dfvs,katara2021deepmpcvs,levine2016end,puang2020kovis,wang2022end} extract 2D keypoints for the subsequent controller. \cite{adrian2022dfbvs,harish2020dfvs,katara2021deepmpcvs} use neural networks to predict matched visual features or optical flow, then calculate control command through IBVS controller. However, IBVS has internal deficiency and may fail to servo the desired pose with large initial pose offset. Other methods\cite{levine2016end,puang2020kovis,wang2022end} use neural controllers instead of IBVS controller. \cite{levine2016end} learn policies that map raw image observations directly to torques at the robot’s motors through deep convolutional neural networks with self-supervised learning. \cite{puang2020kovis} learns the 2D keypoint representation from the image with an auto-encoder and learns the motion based on extracted keypoints. The controllers in the above two methods are trained end-to-end by self-supervised learning. The extracted keypoints can also be used to learn robot motion with end-to-end reinforcement learning \cite{wang2022end}.

\section{METHODS}
HPN-NC generates a unique neural controller for each desired pose in 6 DOF space. In this chapter, we first introduce implementation details of HPN-NC in Section \ref{HPN-NC}. In Section \ref{E2E}, we introduce a three stage training procedure that enables HPN-NC to servo real world objects and to adapt to unseen scenes.

\begin{figure*}[t]
\centering
\includegraphics[width=0.98\linewidth]{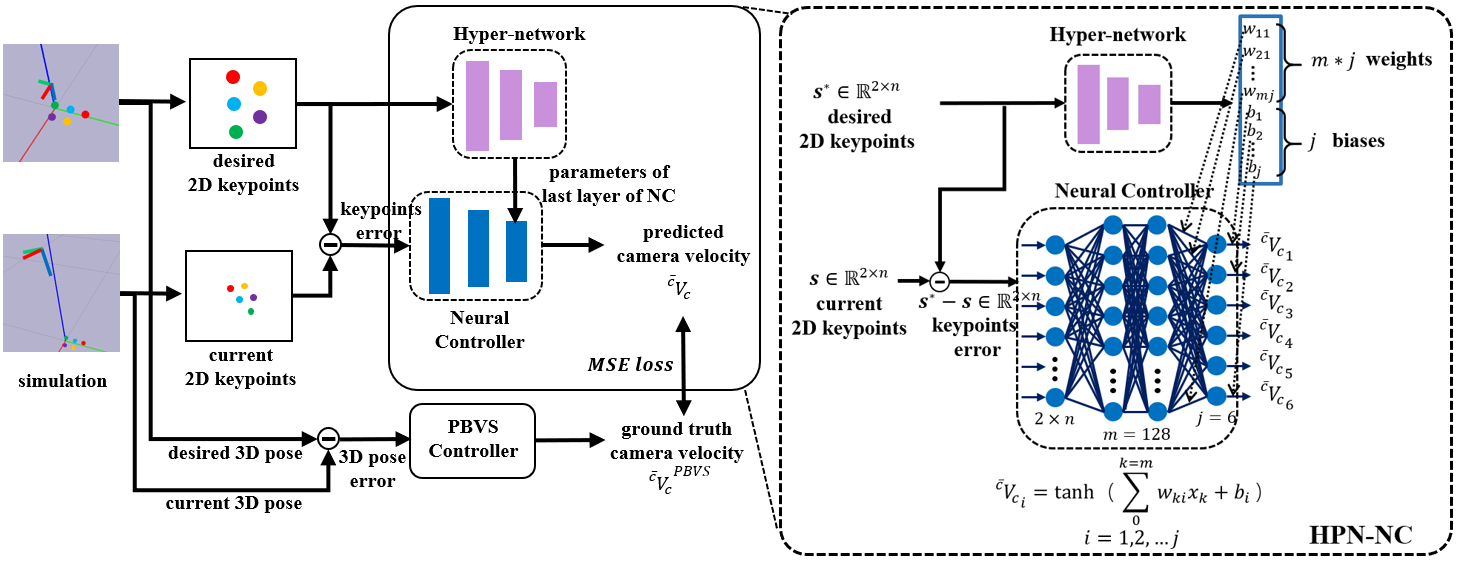}
\caption{The left part gives the pipeline of training HPN-NC in simulation. HPN-NC is supervised by PBVS to ensure a satisfactory servo performance. DAgger helps the model learn more quickly. The right part shows the detail of how HPN generates a NC for a given desired pose. To switch between different desired poses, HPN inferences the weights and biases of the low-level NC's last layer with the 2D keypoints obtained at the desired pose.}
\label{fig:fig2_HPN_NC}
\vspace{-0.55cm}
\end{figure*}

\subsection{Hyper-network Neural Controller}
\label{HPN-NC}
As shown in right part of Fig.~\ref{fig:fig2_HPN_NC}, HPN-NC has a hyper network (HPN, in pink) specialized in encoding the information of desired poses, and a low-level neural controller (NC, in blue) responsible for servoing the given pose. Both of the upper HPN and the low-level NC are lightweighted three-layer fully connected nerual networks. HPN takes the pixel coordinates $\boldsymbol{s}^{*} \in \mathbb{R}^{2 \times n}$ of $n$ 2D keypoints extracted at the desired pose as input and outputs the weights and biases of the last layer of the low-level NC. The last two layers of NC have 128 and 6 units respectively, so HPN outputs 128x6 weights and 6 biases. The low-level NC takes $\boldsymbol{e} = \boldsymbol{s}^{*}-\boldsymbol{s}$ as input just like traditional IBVS controller, which is the coordinate error of the 2D keypoints between the desired pose and the current pose. The output of NC is the 6-dimensional control command ${ }^{\bar{c}} \boldsymbol{V}_{c}=\left[\begin{array}{ll}
v_{c} & \omega_{c}
\end{array}\right]^{T} \in \mathbb{R}^{6}$ consists of instantaneous camera linear velocity and angular velocity under camera frame. Since the last layer parameters of NC are determined by the desired pose, we are able to generate an independent controller for each desired pose and avoid any fine-tuning when switching the desired pose.

 HPN-NC is trained in Pybullet simulation automatically. As shown in left part of Fig.~\ref{fig:fig2_HPN_NC}, like traditional VS, we first set the virtual camera to a random desired pose under the object frame $^{o} \boldsymbol{T}_{c^{*}}$ to get desired 2D keypoints $\boldsymbol{s}^{*}$,
\begin{equation}
    ^{o} \boldsymbol{T}_{c^{*}}=\left[\begin{array}{cc}^{o} \boldsymbol{R}_{c^{*}} & ^{o} \boldsymbol{T}_{c^{*}} \\ 0 & 1\end{array}\right] \in \mathbb{R}^{4 \times 4}
\vspace{-0.1cm} 
\end{equation}
where $o$ represents the object frame and $c^{*}$ represents desired camera frame.
Given the 3D model of the object and the camera intrinsic matrix, we can obtain $\boldsymbol{s}^{*}$ by projection. Taking the 2D keypoint $\boldsymbol{s}^{*}$ of the desired pose as input, HPN infers the parameters of NC's last layer, while the other two layers of NC are identical for any desired pose, 
\begin{equation}
    \boldsymbol{\theta}_{NC}= f^{HPN}_{\boldsymbol{\theta}_{HPN}}\left(\boldsymbol{s}^{*}\right)
\vspace{-0.1cm} \end{equation}
Then, we set the camera to a random initial pose $^{o} \boldsymbol{T}_{c}$, and get current keypoints $\boldsymbol{s}$. The low-level NC takes the keypoint error $\boldsymbol{e} = \boldsymbol{s}^{*}-\boldsymbol{s}$ as input 
and outputs the camera velocity, 
\begin{equation}
    { }^{\bar{c}} \boldsymbol{V}_{c}= f^{NC}_{\boldsymbol{\theta}_{NC}}\left(\boldsymbol{s}^{*}-\boldsymbol{s}\right)=f^{NC}_{\boldsymbol{\theta}_{HPN},\boldsymbol{s}^{*}}\left(\boldsymbol{s}^{*}-\boldsymbol{s}\right)
\vspace{-0.1cm} \end{equation}
PBVS provides the supervision as its trajectory in 3D space is an efficient and secure straight line. We get the supervision through desired pose $^{o} \boldsymbol{T}_{c^{*}}$ and current pose $^{o} \boldsymbol{T}_{c}$:
\begin{equation}
    { }^{\bar{c}} \boldsymbol{V}_{c}^{P B V S}=-\lambda\left[\begin{array}{cc}
{ }^{c^{*}} \boldsymbol{R}_{c}^{T} \cdot{ }^{c^{*}} \boldsymbol{t}_{c} \\
\theta u
\end{array}\right]
\vspace{-0.1cm} \end{equation}
where $\theta u$ is the axial angle of the rotation between current and
the desired pose, $\lambda$ is the coefficient that uniformly set as 0.4 in this work. We denote the input and output tuples to be $q_{NC}$ and the collected training dataset to be $D_{NC}$: 
\begin{equation}
\begin{gathered}
q_{NC} \triangleq (\boldsymbol{s}^{*},\boldsymbol{s},^{\bar{c}} \boldsymbol{V}_{c}^{PBVS}) \\
D_{NC} = \{q_{NC}\}
\end{gathered}
\vspace{-0.1cm} \end{equation}
We use MSE loss $\mathcal{L}_{NC}$ for training and dataset aggregation (DAgger) technique\cite{ross2011reduction} for training acceleration:
\begin{equation}
\begin{gathered}
\mathcal{L}_{NC} =  \|^{\bar{c}} \boldsymbol{V}_{c}^{PBVS}-f^{NC}_{\boldsymbol{\theta}_{HPN},\boldsymbol{s}^{*}}\left(\boldsymbol{s}^{*}-\boldsymbol{s}\right) \|_{2}^{2} 
\end{gathered}
\end{equation}
The upper HPN could be a large and powerful network that has strong encoding ability. The lower NC is a lightweight fully connected network without any complicated structure. For a given desired pose, the parameter inference of NC is done before visual servoing. so the control command inference by NC during VS is efficient. Therefore, HPN-NC takes both the strong modulation for the variation of desired poses and the efficient control inference into account. 

\begin{figure*}[t]
\centering
\includegraphics[width=\linewidth]{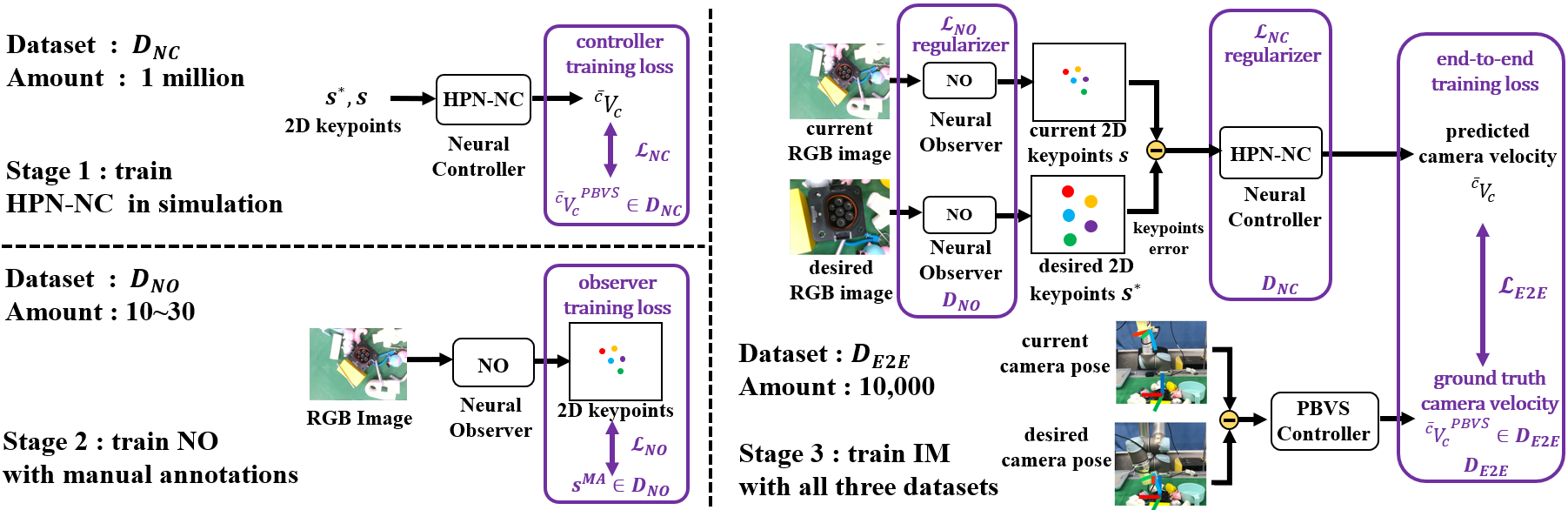}
\caption{The training procedure for real world VS. Image and control tuples are automatically collected to optimize the integrated model consisted of a neural observer and a neural controller. Through end-to-end training, the performance of integrated model could be improved. Note that manual annotations and the simulation data are respectively used as regularizer for observer and controller, so the amount of manual annotations can be significantly reduced.}
\label{fig:fig3_E2E}
\vspace{-0.55cm}
\end{figure*}

\subsection{Real world VS with HPN-NC}
\label{E2E}

We trained a neural observer (NO) to obtain 2D keypoints of ordinary objects in real world (see Fig.~\ref{fig:fig11_3Dmodel}), so that it can be used as the front end of VS controllers. In order to servo objects in a large range in Cartesian space, NO needs to ensure the consistency between 2D keypoints extracted at different poses and the ground truth keypoints projected by the pre-defined 3D model on the camera plane, and also be able to adapt to various backgrounds. We train NO with manual annotated 2D keypoints. To improve its robustness, we expand the dataset with techniques such as translation, rotation, scaling, background replacement and homography matrix stretching. But there still remains some shortcomings: due to the limited manual annotations, it is impossible to cover all the viewing angles in the workspace and NO may fail at certain camera poses, affecting the performance of VS; VS the target object in a new scenario may cause the performance degradation of NO; manual annotating 2D keypoints is costly. 

These shortcomings are fatal for traditional controllers, but not for neural controllers. Being fully differentiable, the integrated model (IM, shown in Fig.~\ref{fig:fig3_E2E}) consists of NO and HPN-NC can be fine-tuned in an end-to-end manner in unseen scenarios. As the supervision is calculated automatically by camera poses, the training process can be self-supervised, which leads to lower data acquisition cost than manual labeling. Therefore, we can utilize a large amount of end-to-end data to improve IM, and use the manual annotations only as the regularizer. This greatly reduces the amount of the manual annotations required for training.


\textbf{Stage1 Training of Controller:}  
The training procedure of HPN-NC is described in Section \ref{HPN-NC}. 

\textbf{Stage2 Training of Observer:} The input of the neural observer is a RGB image $I$, and the output is the pixel coordinates of $n$ 2D keypoints. The backbone of NO can be SpatialConfiguration-Net(SCN)\cite{payer2019integrating} or pre-trained ResNet\cite{he2016deep}. For training dataset $D_{NO}$, we collect RGB images from various perspectives, distances and illuminations, and manually annotate the ground truth 2D keypoint coordinates $\boldsymbol{s}^{MA}_{i}$ of pre-defined 3D model. We have 
\begin{equation}
\begin{gathered}
q_{NO} \triangleq (I,\boldsymbol{s}^{MA}_{i}) \ for \ i=1,2,...,n \\
D_{NO}=\{q_{NO}\}
\end{gathered}
\vspace{-0.1cm} \end{equation}
We use data augmentation techniques described above to improve the generalization of NO. NO generates a heatmap $h_i(x)$ for each keypoint $i$:
\begin{equation}
\begin{gathered}
h_{i}(x) = f^{NO}_{\boldsymbol{\theta}_{NO}}(x) \ for \ i=1,2,...,n \ 
\end{gathered}
\vspace{-0.1cm} \end{equation}
where $x$ is pixel coordinates in $I$.

We tries to minimize the difference between the predicted heatmap and the ground truth heatmap $g_i(x)$ peaking at $\boldsymbol{s}^{MA}$. At the same time, in order to improve the accuracy of predicted keypoints, we minimize the $\boldsymbol{L_2}$ norm between the keypoint coordinates calculated by spatial-softmax operation and the ground truth coordinates $\boldsymbol{s}^{MA}_{i}$. Therefore,the total loss $\mathcal{L}_{NO}$ to train the observer is:
\begin{equation}
\begin{gathered}
\mathcal{L}_{NO} = \sum_{i=1}^{n}(\gamma_{\mathit{h} } \sum_{x \in I}\left\|g_{i}(x)-h_{i}(x)\right\|_{2}^{2} \\
+ \gamma_{\mathit{k} } \left\|\boldsymbol{s}^{MA}_{i}-\sum_{x \in I}x h_{i}(x)\right\|_{2}^{2}) \\
\end{gathered}
\label{equ:loss_NO}
\vspace{-0.1cm} 
\end{equation}
where $\gamma_{\mathit{h} }=10$ and $\gamma_{\mathit{k} }=0.00001$ are scale factors to facilitate the learning process convergence.

\textbf{Stage3 End-to-end Training of Integrated Model:} We fine-tune IM with large amount robot's self-supervision data in an end-to-end manner. As shown in Fig.~\ref{fig:fig3_E2E}, a random desired pose under the robot's base frame $^{b} \boldsymbol{T}_{c^{*}}$ is sampled, the robot first moves the camera to this pose. An image $I^*$ observed at the desired poses $^{b} \boldsymbol{T}_{c^{*}}$ is sent to NO to obtain desired 2D keypoints $\boldsymbol{s}^{*}$. Afterwards the robot moves the camera to a random sampled initial pose $^{b} \boldsymbol{T}_{c}$ to get the initial 2D keypoints $\boldsymbol{s}$,
\begin{equation}
\boldsymbol{s}^{*}=f^{NO}_{\boldsymbol{\theta}_{NO}}\left(I^{*}\right), \boldsymbol{s}=f^{NO}_{\boldsymbol{\theta}_{NO}}(I)
\vspace{-0.1cm} \end{equation}
HPN infers an unique NC for $^{b} \boldsymbol{T}_{c^{*}}$ according to $\boldsymbol{s}^{*}$,
\begin{equation}
\boldsymbol{\theta}_{NC}=f^{HPN}_{\boldsymbol{\theta}_{HPN}}\left(\boldsymbol{s}^{*}\right)
\vspace{-0.1cm} \end{equation}
NC outputs the velocity of the camera according to the 2D keypoint error $\boldsymbol{e}=\boldsymbol{s}^{*}-\boldsymbol{s}$,
\begin{equation}
{ }^{\bar{c}} \boldsymbol{V}_{c}=f^{NC}_{\boldsymbol{\theta}_{HPN},\boldsymbol{s}^{*}}\left(\boldsymbol{s}^{*}-\boldsymbol{s}\right)
\vspace{-0.1cm} \end{equation}
We use DAgger to deal the out of distribution problem. When the robot is wandering in the workspace following ${ }^{\bar{c}} \boldsymbol{V}_{c}$, it automatically collects the end-to-end (E2E) training dataset $D_{E2E}$. $D_{E2E}$ is consist of image and control tuple $q_{E2E}$ at different poses:
\begin{equation}
\begin{gathered}
q_{E2E} \triangleq (I,I^{*},{ }^{\bar{c}} \boldsymbol{V}_{c}^{PBVS}) \\
D_{E2E}=\{q_{E2E}\} \\
\end{gathered}
\vspace{-0.1cm} \end{equation}
where ${ }^{\bar{c}} \boldsymbol{V}_{c}^{PBVS}$ is calculated by PBVS controller according to desired pose $^{b} \boldsymbol{T}_{c^{*}}$ and the current pose $^{b} \boldsymbol{T}_{c}$:
\begin{equation}
{ }^{\bar{c}} \boldsymbol{V}_{c}^{PBVS}=-\lambda\left[\begin{array}{c}
{ }^{c^{*}} \boldsymbol{R}_{c}^{T} \cdot{ }^{c^{*}} \boldsymbol{t}_{c} \\
\theta u
\end{array}\right]
\vspace{-0.1cm} \end{equation}
We use MSE loss $\mathcal{L}_{E2E}$ for training:
\begin{equation}
\begin{gathered}
\mathcal{L}_{E2E} =  \left\|^{\bar{c}} \boldsymbol{V}_{c}^{PBVS}-f^{NC}_{\boldsymbol{\theta}_{HPN},I^{*}}\left(f^{NO}_{\boldsymbol{\theta}_{NO}}\left(I^{*}\right)-f^{NO}_{\boldsymbol{\theta}_{NO}}(I)\right)\right\|_{2}^{2} \\
\end{gathered}
\vspace{-0.1cm} \end{equation}
However, there are actually infinite kinds of IMs that satisfy the constraints from $D_{E2E}$, results in a drift in the outputs of observer NO and controller HPN-NC. To prevent drift, we want NO and HPN-NC to satisfy the constraints from $D_{NO}$ and $D_{NC}$. Thus, we use $D_{NO}$ and $D_{NC}$ to co-train NO and HPN-NC with $D_{E2E}$, so the data in $D_{NO}$ and $D_{NC}$ will regularize NO and HPN-NC. Since the data in $D_{NO}$ only acts as the regularizer, only a small number of manual annotations is needed. The total loss function of Stage 3 is:
\begin{equation}
\begin{gathered}
\mathcal{L} = \mathcal{L}_{NC} + \mathcal{L}_{NO} + \mathcal{L}_{E2E}
\end{gathered}
\vspace{-0.1cm} \end{equation}
Note that when a calibrated camera extrinsic matrix is given, the robotic arm can move the camera to a specified pose automatically, also the end-to-end supervision does not require any manual labeling, so the entire learning process can be fully automated. 

\section{SYSTEM IMPLEMENTATION}

\subsection{Simulation Settings}
\label{Simulation Settings}
An environment including a virtual camera and the target object's 3D model is built in Pybullet. In each data collecting or model evaluation episode, a random desired camera pose $^{o} \boldsymbol{T}_{c^{*}}$ is sampled in the space 15cm above the target object with 0 to 5cm disturbance in $XYZ$ translation and a random initial camera pose $^{o} \boldsymbol{T}_{c}$ is sampled in the space 30cm above the target object with 0 to 10cm disturbance in $XYZ$ translation. Both $^{o} \boldsymbol{T}_{c}$ and $^{o} \boldsymbol{T}_{c^{*}}$ ensure all keypoints within the camera's field of view (Fov). The maximum initial pose offset between the initial and desired pose is $\Delta \mathbf{r}_{0}=(^{{c}^*}\mathbf{t}_{c},\mathbf{\theta}\mathbf{u}):$ $^{{c}^*}\mathbf{t}_{c} = (15cm,15cm,30cm), \mathbf{\theta}\mathbf{u} = (53.1\degree,53.1\degree,180\degree,)$. An episode is considered to be finished successfully if the total error between the current  and the desired keypoints is lower than a specified threshold $\delta_f$:
\begin{equation}
    \sum _{k=1} ^n \left| u_k - u_k^* \right| + \left| v_k - v_k^* \right| \leq \delta_f
\vspace{-0.1cm} \end{equation}
Before the keypoint error fully converges, there are several situations that trigger the early termination of the episode:
\begin{itemize}
\item Every episode has a maximum steps of $\delta_s$ with each step takes $0.1s$. An episode will finish if the keypoint error hasn't reached the threshold $\delta_f$ within $\delta_s$ steps.
\item The camera walks out the workspace.
\item Any 2D keypoint is out of the camera's Fov.
\end{itemize}

\begin{figure}[t]
\centering
\includegraphics[width=\linewidth]{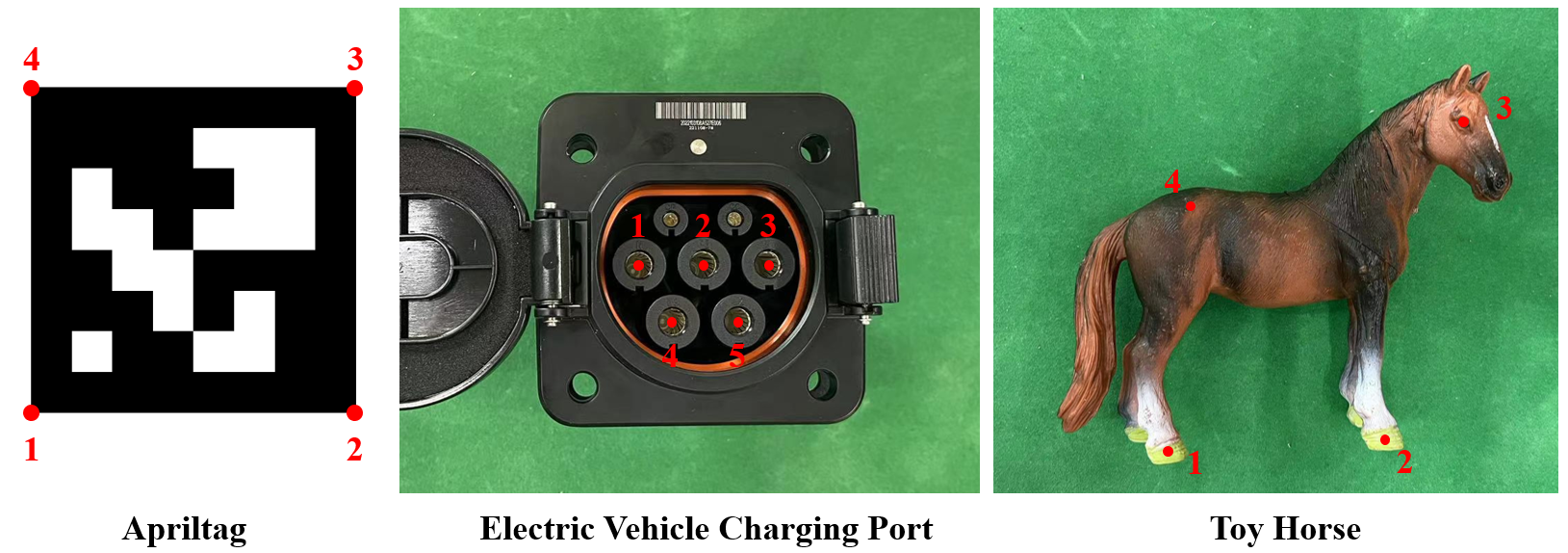}
\caption{Target objects: Apriltag is used for evaluation of HPN-NC in simulation. The charging port and the toy horse are real world objects without obvious feature. Specific values of 3D models is in the supplementary materials.}
\label{fig:fig11_3Dmodel}
\vspace{-0.5cm}
\end{figure}

We use four criterias to analysis the servo performance: servo success rate (\textbf{SR}), servo efficiency (timesteps,\textbf{TS}), final rotation error (\textbf{RE}) and final translation error (\textbf{TE}). We calculate the transformation between the final camera pose and the desired pose $^{{c}^*}\boldsymbol{T}_{c}$.

\textbf{Rotation Error (RE):} The relative rotation $^{{c}^*}\boldsymbol{R}_{c}$ is converted into an axis-angle representation $^{{c}^*}\theta_{c} \ ^{{c}^*}\boldsymbol{u}_{c}$. The rotation accuracy of camera is considered satisfactory if the deflection angle between the final pose and the desired pose is less than threshold $\delta_r$.

\textbf{Translation Error (TE):} The translation accuracy of camera is considered satisfactory if the displacement between the final position and the desired position is less than $\delta_t$.

If RE and TE are less than corresponding threshold, an episode would also be regarded as a successful trial. For threshold values, please refer to the supplementary material.

\subsection{Real World Settings}
The real world experiment is carried out on an UR5 robot. The settings of the real world experiment are the same as those of the simulation except for some thresholds. For specific parameter values of real world experiments, please refer to the supplementary material.

\subsection{Target Objects and 3D Models}

Compared with some pioneer works\cite{puang2020kovis,adrian2022dfbvs} that need accurate objects' meshes for rendering and training, we only need to define $n$ 3D feature points on the object and roughly measure their coordinates under object's frame as 3D model. 
Fig.~\ref{fig:fig11_3Dmodel} shows several objects we used as 
 target objects. For specific coordinates of their 3D models, please refer to the supplementary material.
 

\section{EXPERIMENTAL RESULTS}
In this section, we carried out a series of experiments to evaluate our method. Simulation experiments are performed on a computer with 16 Intel(R) Core(TM) i9-9900K 3.60GHz and one NVIDIA GeForce RTX 2080 SUPER. Real world experiments are performed on a computer with 12 Intel(R) Core(TM) i7-8700 3.20GHz and one NVIDIA GeForce GTX 1060. The goals of the experiments are:
\begin{itemize}
    \item to validate that the performance of proposed HPN-NC is better than the traditional IBVS controller and other neural controllers in multiple desired poses VS tasks.
    \item to validate that the integrated model (IM) is able to servo real world object with no obvious feature in unseen scenes.
    \item to validate that IM can further improve the servo performance in unseen scene, promote the keypoint extraction ability and  reduce manual annotation cost with self-supervised end-to-end training.
\end{itemize}

\subsection{Introduction of Baseline} 
\label{Simulation Evaluations}
We compare the performance of HPN-NC with IBVS\cite{chaumette2006basic} and three neural controllers. These NCs have different structures but are all supervised by the same PBVS teacher. Fully connected neural controller (FCN-NC) is a three layers fully connected neural networks derived from \cite{levine2016end}, whose input is the 2D keypoints error of current and desired pose. DenseNet based neural controller (DenseNet-NC) is the controller used in \cite{puang2020kovis}. DenseNet-NC's main structure is a DenseNet following with a fully connected layer. It takes the concatenated vector of current and desired keypoints as input. Auto-encoder based neural controller (AE-NC) draws on the structure of auto-encoder\cite{worrall2017interpretable} to encode the information of the desired pose. Its low-level controller is similar to FCN-NC except for an additional input: a latent vector from the auto-encoder. For implementation details, please refer to the supplementary material. 

\subsection{Simulation Evaluations} 
\label{Simulation Evaluations}

\begin{figure}[htbp]
\centering
\includegraphics[width=0.99\linewidth]{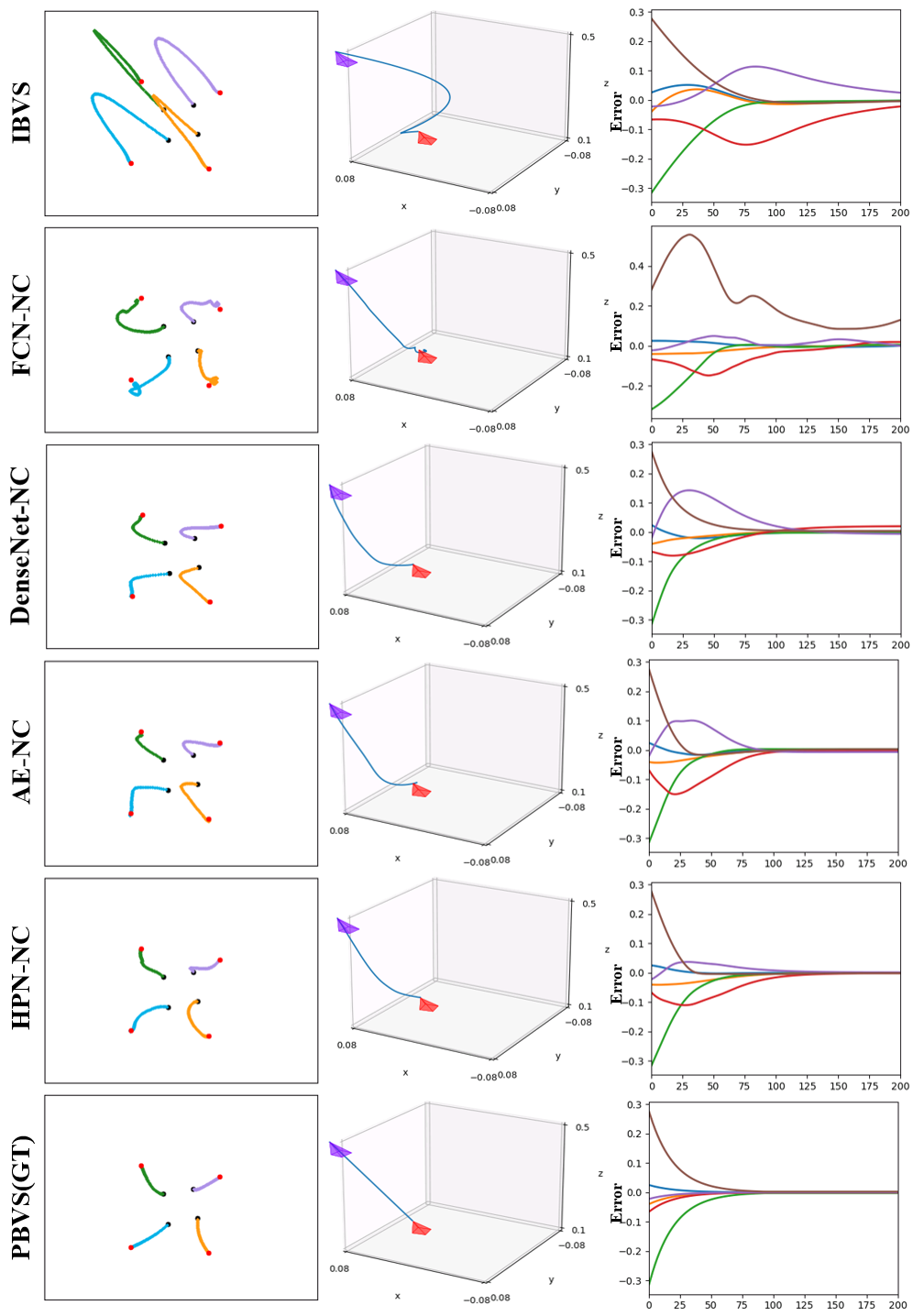}
\caption{2D, 3D and error trajectories for a VS task. 
For 2D trajectories, black dots represent the initial keypoints and red dots represent the desired keypoints. For 3D trajectories, purple triangles represent the initial camera poses and red triangles represent the desired camera poses. Error trajectories visualize the TE and RE between the current and the desired camera poses. TE is in meters and RE is represented by the axis angle.}
\label{fig:fig16_case1}
\vspace{-0.5cm}
\end{figure}

\textbf{Performance Comparison:}
All the NCs are trained for 100 epochs. Each epoch first runs 10 thousands data collection steps. Then NCs are trained for 500 batches with a batch size of 512. The evaluation is performed on the apriltag shown in Fig.~\ref{fig:fig11_3Dmodel}. Tab. \ref{table2_simulation_nc} shows the performance of controllers for 500 trials with different desired poses. Supervised by a perfect PBVS teacher, all of NCs have higher SR than IBVS. Among these NCs, HPN-NC has higher SR, servo accuracy (RE and TE) and shorter inference time. We believe that the advantage of HPN-NC stems from the fact that it uses a complex hyper network to model the modulate mechanism of different VS tasks, but only the lightweight low-level fully connected NC is used during the servo process, which ensures efficient inference. Fig. \ref{fig:fig16_case1} shows the 2D, 3D and error trajectories for two visual servoing tasks with different desired and initial poses. All controllers have reached the desired pose. But only DenseNet-NC, AE-NC and HPN-NC complete within 200 steps. The terminal error of HPN-NC is smaller than that of DenseNet-NC and AE-NC because of stronger modulate ability. With such strong hyper net, HPN-NC's 2D and 3D trajectories are more similar to the ground truth PBVS's. For more cases, please refer to the supplementary material.

\begin{figure}[htbp]
\centering
\includegraphics[width=0.67\linewidth]{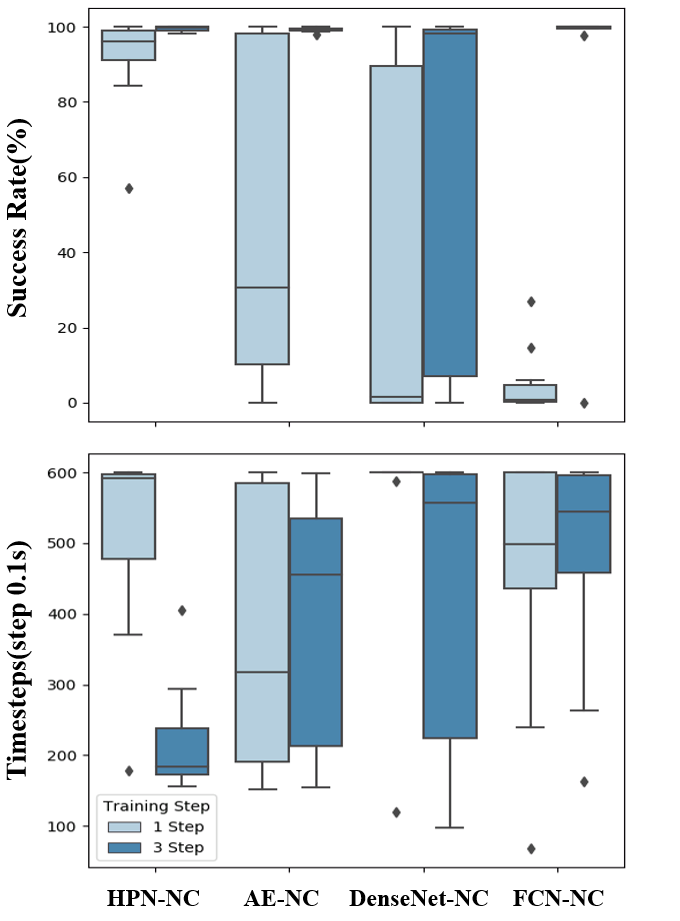}
\caption{We select 10 desired poses that all the NCs in Tab. \ref{table2_simulation_nc} failed and compare the performance of fine-tuned NCs after 1-step and 3-step fine-tuning.}
\label{fig:fig7_adapte}
\vspace{-0.5cm}
\end{figure}

\textbf{Fine-tuning with Self-supervised End-to-end Training:} No matter how powerful the neural controllers are, they cannot guarantee that all desired poses can be successfully VS. But since these NCs are completely differentiable, for those desired poses that cannot be servoed, NCs can be fine-tuned through self-supervised end-to-end training. The fine-tuning process is similar to the training process described in Section \ref{HPN-NC}, except that the desired pose is fixed. We select 10 desired poses that all of the NCs in Tab. \ref{table2_simulation_nc} failed to VS and compare the performance of fine-tuned NCs. As shown in Fig.~\ref{fig:fig7_adapte}, we compare the average SR and TS after 1-step and 3-step fine-tuning for 10 desired poses. Each step of fine-tuning runs 10 thousand data collection steps. Then the model is fine-tuned for 500 batches with a batch size of 512. For evaluation, the fine-tuned models try to VS the selected desired poses from 500 different initial poses. After 1-step fine-tuning, HPN-NC’s average servo SR is about 90$\%$ for 10 desired poses, which is the highest among all NCs’. After 3-step fine-tuning, HPN-NC, AE-NC and FCN-NC all have high SR, but HPN-NC has less TS for about 200 steps which means it can reach the desired poses faster. In other words, only HPN-NC can achieve high success rate and high efficiency VS with efficient fine-tuning.

\subsection{Real World Evaluations}
\label{threestage}

\textbf{Performance Comparison:} VS real world objects in unseen scenes inevitably needs to face the recognition error of the observer. We first compare the performance of different controllers given the same neural observer. We train the NO with SpatialConfiguration-Net\cite{payer2019integrating} (SCN) to extract the pre-defined 2D keypoints of the charging port on 1000 annotations. For specific training details, please refer to the supplementary material. 2D keypoints predicted by NO are used as the input of controllers. Tab.~\ref{table4_IM} shows the results of different integrated models to servo 50 different desired poses. As discussed in Section \ref{E2E}, limited annotations and unseen scene leads to the recognition error of NO. Although affected by recognition error, HPN-NC has the highest SR and smallest TS 
compared with IBVS and other NCs.
The same degradation is happened to PBVS. In Row 7, the SR of PBVS which uses relative camera pose estimated by Perspective-n-Point\cite{li2012robust} (PnP) with 2D keypoints extracted by NO, drops 10$\%$ compared with ground truth PBVS. The relative camera pose of ground truth PBVS is calculated by robot's tool center point and calibrated camera extrinsic, so it will not be affected by NO's recognition error.

\textbf{Performance Improvement with Self-supervised End-to-end Training:} Unfortunately, PnP is not differentiable, IM consists of NO and PBVS cannot be further promoted to deal with recognition error. Taking advantage of the fully differentiable nature of HPN-NC, IM consists of NO and HPN-NC can be improved by self-supervised end-to-end training with DAgger. IM is fine-tuned for 5 epochs. Each epoch runs 2000 data collection steps to get $D_{E2E}$. $D_{NC}$, $D_{NO}$ and $D_{E2E}$ are divided to be the training set with 80$\%$ data and the validation set with 20$\%$ data. Then, IM is fine-tuned for 2000 epochs with a batch size of 512,2,1 respectively for $D_{NC}$, $D_{NO}$ and $D_{E2E}$. Lastly, a model with smallest end-to-end loss is selected for the next DAgger epoch. As shown in Tab.~\ref{table4_IM}, end-to-end training improves IM's SR from 78$\%$ to 98$\%$. Servo efficiency(TS) and accuracy(RE and TE) are also improved. As discussed in Stage 3 of Section \ref{E2E}, during the end-to-end training, the robot is self-supervised and no manual effort is introduced.

\begin{figure}[t]
\centering
\includegraphics[width=0.6\linewidth]{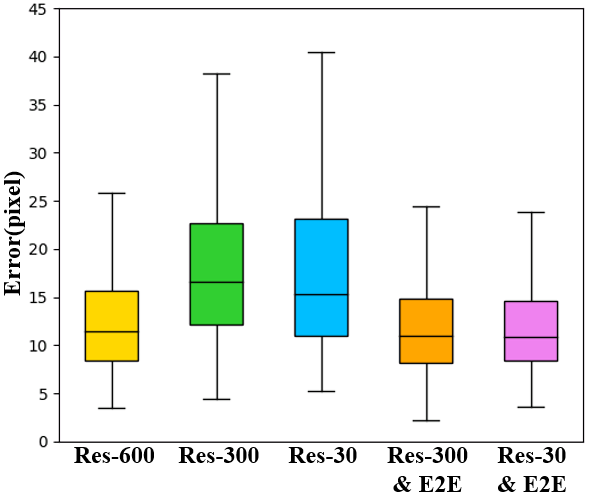}
\caption{The total recognition error of five 2D keypoints for the charging port. Less training data for NO introduces larger recognition error. With self-supervised end-to-end training, NO trained with 30 pieces manual annotations can reduce the recognition error to 600 piece's level.}
\label{fig:fig15_NO_prediction_error3}
\vspace{-0.48cm}
\end{figure}

\textbf{Manual Annotation Reduction with Self-supervised End-to-end Training:} A more realistic problem is that real world objects' annotations are often insufficient due to the high production costs. Less training data introduces larger recognition error which causes VS performance degradation. Another advantage of self-supervised end-to-end training is to reduce the amount of manual annotations needed for VS. By using those annotations only as regularizer, we replace the expensive manual annotations with cheap self-supervised end-to-end data with control labels. We choose a pre-trained ResNet-18\cite{krizhevsky2012imagenet} as NO to avoid the failure of training NO with too little annotations. We respectively use 600, 300 and 30 pieces of annotations to train NO. As shown in Tab.~\ref{table4_MAR}, SR of IM with PBVS(PnP) gradually decreases as the amount of manual annotations decreases. Through the end-to-end training described in Section \ref{E2E}, SR of IMs can respectively be promoted to 94$\%$(300 annotations) and 92$\%$(30 annotations). From Fig.~\ref{fig:fig15_NO_prediction_error3}, we could find the recognition error can be reduced after end-to-end training.

\begin{table*}[t]
\centering
\caption{Performance comparison of controllers in simulation. The bold face represents the best. The underline represents the second best. To avoid failed trials to interfere with the statistic results, TS, RE and TE only count those successful trials.} 
\label{table2_simulation_nc}
\begin{tabular}{c|c|c|c|c|c|c}
\hline
\hline
\textbf{Controller} & \textbf{SR(\%)} & \textbf{TS(0.1s)} & \textbf{RE(rad)} & \textbf{TE(cm)} & \textbf{$\#$Parameters} & \textbf{Inference time(ms)} \\ \hline

IBVS\cite{chaumette2006basic}                & 17.2                      & \bf{324.39$\pm$95.66} & 0.039$\pm$0.036           & 0.861$\pm$0.835                      & \textbackslash{}        & \textbackslash{}           \\ \hline
FCN-NC\cite{levine2016end}                   & 27.8                      & 510.57$\pm$156.36     & 0.029$\pm$0.014           & 0.494$\pm$0.197            & \bf{69638}              & \underline{0.160}           \\ \hline
DenseNet-NC\cite{puang2020kovis}             & 87                        & 523.42$\pm$164.74     & 0.020$\pm$0.009           & 0.452$\pm$0.202            & 281558                  & 1.320                   \\ \hline
AE-NC\cite{worrall2017interpretable}         & \underline{89.8}          & \underline{441.59$\pm$210.74}    & \underline{0.012$\pm$0.008}         & \underline{0.327$\pm$0.192}          & \underline{71686}       & 0.164                   \\ \hline
\bf{HPN-NC(Ours)}                            & \bf{95}                   & 443.9$\pm$205.33      & \bf{0.005$\pm$0.006}      & \bf{0.193$\pm$0.119}       & \bf{69638(532486)}      & \bf{0.158}                   \\ \hline
\hline
\end{tabular}
\vspace{-0.37cm}
\end{table*}

\begin{table}[t]
\centering
\caption{Real world comparison of different controllers with the same NO in unseen scene. The bold face represents the best.}
\label{table4_IM}
\resizebox{\linewidth}{!}
{\begin{tabular}{c|c|c|c|c}
\hline
\hline
\textbf{Controller}& \textbf{SR(\%)} & \textbf{TS(0.1s)} & \textbf{RE(rad)} & \textbf{TE(cm)} \\ \hline
PBVS(GT)                            & 100                & 60.76$\pm$26.39          & 0.022$\pm$0.008           & 0.299$\pm$0.180         \\ \hline
IBVS\cite{chaumette2006basic}       & 30                 & 101.20$\pm$39.70         & 0.221$\pm$0.194           & 4.942$\pm$4.631         \\ \hline
FCN-NC\cite{levine2016end}          & 34                 & 128.31$\pm$65.73         & 0.155$\pm$0.126           & 3.263$\pm$2.745         \\ \hline
DenseNet-NC\cite{puang2020kovis}    & 58                 & 120.14$\pm$62.78         & 0.113$\pm$0.072           & 2.471$\pm$1.603         \\ \hline
AE-NC\cite{worrall2017interpretable}& 70                 & 94.38$\pm$49.22          & 0.128$\pm$0.160           & 3.084$\pm$3.513         \\ \hline
HPN-NC                              & 78                 & 93.00$\pm$47.32          & 0.132$\pm$0.109           & 2.902$\pm$2.447         \\ \hline
PBVS(PnP)\cite{chaumette2006basic}  & 90                 & \bf{60.89$\pm$23.58}     & 0.137$\pm$0.120           & 2.961$\pm$2.672         \\ \hline
\bf{HPN-NC$\&$E2E}                  & \bf{98}            & 90.16$\pm$56.34          & \bf{0.051$\pm$0.024}      & \bf{1.113$\pm$0.519}    \\ 
\hline
\hline 
\end{tabular}}
\vspace{-0.2cm}
\end{table}

\begin{table}[t]
\centering
\caption{Real world comparison of IMs trained with different amount of manual annotations. The bold face represents the best.}
\label{table4_MAR}
\resizebox{\linewidth}{!}
{\begin{tabular}{l|c|c|c|c}
\hline
\hline
\textbf{Integrated Model}& \textbf{SR(\%)}  & \textbf{TS(0.1s)}     & \textbf{RE(rad)}   & \textbf{TE(cm)}     \\ \hline
Res-600 PBVS(PnP)           & 88            & 90.29$\pm$34.60       & 0.142$\pm$0.133    & 2.950$\pm$2.865     \\ \hline
Res-300 PBVS(PnP)           & 56            & 92.00$\pm$32.58       & 0.115$\pm$0.201    & 2.391$\pm$4.171     \\ \hline
Res-30 PBVS(PnP)            & 30            & \bf{89.06$\pm$25.90}  & 0.266$\pm$0.307    & 4.365$\pm$3.928     \\ \hline
Res-300 HPN-NC$\&$E2E     & \bf{94}       & 119.34$\pm$53.27      & 0.067$\pm$0.038    & 1.431$\pm$0.778     \\ \hline
Res-30 HPN-NC$\&$E2E      & 92            & 112.48$\pm$44.05      &\bf{0.059$\pm$0.036}&\bf{1.275}$\pm$0.745 \\ \hline
\hline 
\end{tabular}}
\vspace{-0.3cm}
\end{table}

\section{CONCLUSIONS}

In this paper, we explore that hyper-network is an appropriate architecture for multiple desired poses VS. It outperforms IBVS and other neural controllers by success rate, servo efficiency, network volume, inference time and adaptation efficiency. We evaluate the proposed model in both simulation and real world experiments. For real world VS task, we propose a three-stage training procedure that can further improve the model’s servo performance and reduces manual annotation amount. It’s fully automatic and achieves 92$\%$ success rate with only 30 pieces of manual annotations. With the proposed training procedure, VS can be efficiently applied to similar scenarios in real world. In the future, we will address the task with more matched correspondences such as eye-to-hand VS with optical flow.









\bibliographystyle{IEEEtran}
\bibliography{root}

\begin{thebibliography}{10}
\providecommand{\url}[1]{#1}
\csname url@samestyle\endcsname
\providecommand{\newblock}{\relax}
\providecommand{\bibinfo}[2]{#2}
\providecommand{\BIBentrySTDinterwordspacing}{\spaceskip=0pt\relax}
\providecommand{\BIBentryALTinterwordstretchfactor}{4}
\providecommand{\BIBentryALTinterwordspacing}{\spaceskip=\fontdimen2\font plus
\BIBentryALTinterwordstretchfactor\fontdimen3\font minus
  \fontdimen4\font\relax}
\providecommand{\BIBforeignlanguage}[2]{{%
\expandafter\ifx\csname l@#1\endcsname\relax
\typeout{** WARNING: IEEEtran.bst: No hyphenation pattern has been}%
\typeout{** loaded for the language `#1'. Using the pattern for}%
\typeout{** the default language instead.}%
\else
\language=\csname l@#1\endcsname
\fi
#2}}
\providecommand{\BIBdecl}{\relax}
\BIBdecl

\bibitem{chaumette2006basic}
F.~Chaumette and S.~Hutchinson, ``Visual servo control. i. basic approaches,''
  \emph{IEEE Robotics Automation Magazine}, vol.~13, no.~4, pp. 82--90, 2006.

\bibitem{bakthavatchalam2013photometric}
M.~Bakthavatchalam, F.~Chaumette, and E.~Marchand, ``Photometric moments: New
  promising candidates for visual servoing,'' in \emph{2013 IEEE International
  Conference on Robotics and Automation}.\hskip 1em plus 0.5em minus
  0.4em\relax IEEE, 2013, pp. 5241--5246.

\bibitem{shi2018adaptive}
H.~Shi, G.~Sun, Y.~Wang, and K.-S. Hwang, ``Adaptive image-based visual
  servoing with temporary loss of the visual signal,'' \emph{IEEE Transactions
  on Industrial Informatics}, vol.~15, no.~4, pp. 1956--1965, 2018.

\bibitem{wang2018adaptive}
H.~Wang, B.~Yang, J.~Wang, X.~Liang, W.~Chen, and Y.-H. Liu, ``Adaptive visual
  servoing of contour features,'' \emph{IEEE/ASME Transactions on
  Mechatronics}, vol.~23, no.~2, pp. 811--822, 2018.

\bibitem{saxena2017exploring}
A.~Saxena, H.~Pandya, G.~Kumar, A.~Gaud, and K.~M. Krishna, ``Exploring
  convolutional networks for end-to-end visual servoing,'' in \emph{2017 IEEE
  International Conference on Robotics and Automation (ICRA)}.\hskip 1em plus
  0.5em minus 0.4em\relax IEEE, 2017, pp. 3817--3823.

\bibitem{bateux2018icra}
Q.~Bateux, E.~Marchand, J.~Leitner, F.~Chaumette, and P.~Corke, ``Training deep
  neural networks for visual servoing,'' in \emph{2018 IEEE International
  Conference on Robotics and Automation (ICRA)}, 2018, pp. 3307--3314.

\bibitem{yu2019siamese}
C.~Yu, Z.~Cai, H.~Pham, and Q.-C. Pham, ``Siamese convolutional neural network
  for sub-millimeter-accurate camera pose estimation and visual servoing,'' in
  \emph{2019 IEEE/RSJ International Conference on Intelligent Robots and
  Systems (IROS)}.\hskip 1em plus 0.5em minus 0.4em\relax IEEE, 2019, pp.
  935--941.

\bibitem{felton2021siame}
S.~Felton, E.~Fromont, and E.~Marchand, ``Siame-se (3): regression in se (3)
  for end-to-end visual servoing,'' in \emph{2021 IEEE International Conference
  on Robotics and Automation (ICRA)}.\hskip 1em plus 0.5em minus 0.4em\relax
  IEEE, 2021, pp. 14\,454--14\,460.

\bibitem{adrian2022dfbvs}
N.~Adrian, V.-T. Do, and Q.-C. Pham, ``Dfbvs: Deep feature-based visual
  servo,'' \emph{arXiv preprint arXiv:2201.08046}, 2022.

\bibitem{harish2020dfvs}
Y.~Harish, H.~Pandya, A.~Gaud, S.~Terupally, S.~Shankar, and K.~M. Krishna,
  ``Dfvs: Deep flow guided scene agnostic image based visual servoing,'' in
  \emph{2020 IEEE International Conference on Robotics and Automation
  (ICRA)}.\hskip 1em plus 0.5em minus 0.4em\relax IEEE, 2020, pp. 9000--9006.

\bibitem{katara2021deepmpcvs}
P.~Katara, Y.~Harish, H.~Pandya, A.~Gupta, A.~Sanchawala, G.~Kumar,
  B.~Bhowmick, and M.~Krishna, ``Deepmpcvs: Deep model predictive control for
  visual servoing,'' in \emph{Conference on Robot Learning}.\hskip 1em plus
  0.5em minus 0.4em\relax PMLR, 2021, pp. 2006--2015.

\bibitem{espiau1992new}
B.~Espiau, F.~Chaumette, and P.~Rives, ``A new approach to visual servoing in
  robotics,'' \emph{ieee Transactions on Robotics and Automation}, vol.~8,
  no.~3, pp. 313--326, 1992.

\bibitem{kelly2000stable}
R.~Kelly, R.~Carelli, O.~Nasisi, B.~Kuchen, and F.~Reyes, ``Stable visual
  servoing of camera-in-hand robotic systems,'' \emph{IEEE/ASME transactions on
  mechatronics}, vol.~5, no.~1, pp. 39--48, 2000.

\bibitem{puang2020kovis}
E.~Y. Puang, K.~P. Tee, and W.~Jing, ``Kovis: Keypoint-based visual servoing
  with zero-shot sim-to-real transfer for robotics manipulation,'' in
  \emph{2020 IEEE/RSJ International Conference on Intelligent Robots and
  Systems (IROS)}.\hskip 1em plus 0.5em minus 0.4em\relax IEEE, 2020, pp.
  7527--7533.

\bibitem{wang2022end}
T.~Wang, E.~Y. Puang, M.~Lee, Y.~Wu, and W.~Jing, ``End-to-end reinforcement
  learning of robotic manipulation with robust keypoints representation,''
  \emph{arXiv preprint arXiv:2202.06027}, 2022.

\bibitem{levine2016end}
S.~Levine, C.~Finn, T.~Darrell, and P.~Abbeel, ``End-to-end training of deep
  visuomotor policies,'' \emph{The Journal of Machine Learning Research},
  vol.~17, no.~1, pp. 1334--1373, 2016.

\bibitem{ha2016hypernetworks}
D.~Ha, A.~Dai, and Q.~V. Le, ``Hypernetworks,'' \emph{arXiv preprint
  arXiv:1609.09106}, 2016.

\bibitem{allibert2010predictive}
G.~Allibert, E.~Courtial, and F.~Chaumette, ``Predictive control for
  constrained image-based visual servoing,'' \emph{IEEE Transactions on
  Robotics}, vol.~26, no.~5, pp. 933--939, 2010.

\bibitem{thuilot2002pbvs}
B.~Thuilot, P.~Martinet, L.~Cordesses, and J.~Gallice, ``Position based visual
  servoing: keeping the object in the field of vision,'' in \emph{Proceedings
  2002 IEEE International Conference on Robotics and Automation (Cat.
  No.02CH37292)}, vol.~2, 2002, pp. 1624--1629 vol.2.

\bibitem{park2012pbvs}
D.-H. Park, J.-H. Kwon, and I.-J. Ha, ``Novel position-based visual servoing
  approach to robust global stability under field-of-view constraint,''
  \emph{IEEE Transactions on Industrial Electronics}, vol.~59, no.~12, pp.
  4735--4752, 2012.

\bibitem{gans2007switch}
N.~R. Gans and S.~A. Hutchinson, ``Stable visual servoing through hybrid
  switched-system control,'' \emph{IEEE Transactions on Robotics}, vol.~23,
  no.~3, pp. 530--540, 2007.

\bibitem{hafez2007weighted}
A.~H.~A. Hafez and C.~Jawahar, ``Visual servoing by optimization of a 2d/3d
  hybrid objective function,'' in \emph{Proceedings 2007 IEEE International
  Conference on Robotics and Automation}, 2007, pp. 1691--1696.

\bibitem{jin2021policy}
Z.~Jin, J.~Wu, A.~Liu, W.-A. Zhang, and L.~Yu, ``Policy-based deep
  reinforcement learning for visual servoing control of mobile robots with
  visibility constraints,'' \emph{IEEE Transactions on Industrial Electronics},
  vol.~69, no.~2, pp. 1898--1908, 2021.

\bibitem{krizhevsky2012imagenet}
A.~Krizhevsky, I.~Sutskever, and G.~E. Hinton, ``Imagenet classification with
  deep convolutional neural networks,'' \emph{Advances in neural information
  processing systems}, vol.~25, 2012.

\bibitem{huang2017densely}
G.~Huang, Z.~Liu, L.~Van Der~Maaten, and K.~Q. Weinberger, ``Densely connected
  convolutional networks,'' in \emph{Proceedings of the IEEE conference on
  computer vision and pattern recognition}, 2017, pp. 4700--4708.

\bibitem{ross2011reduction}
S.~Ross, G.~Gordon, and D.~Bagnell, ``A reduction of imitation learning and
  structured prediction to no-regret online learning,'' in \emph{Proceedings of
  the fourteenth international conference on artificial intelligence and
  statistics}.\hskip 1em plus 0.5em minus 0.4em\relax JMLR Workshop and
  Conference Proceedings, 2011, pp. 627--635.

\bibitem{payer2019integrating}
C.~Payer, D.~{\v{S}}tern, H.~Bischof, and M.~Urschler, ``Integrating spatial
  configuration into heatmap regression based cnns for landmark localization,''
  \emph{Medical image analysis}, vol.~54, pp. 207--219, 2019.

\bibitem{he2016deep}
K.~He, X.~Zhang, S.~Ren, and J.~Sun, ``Deep residual learning for image
  recognition,'' in \emph{Proceedings of the IEEE conference on computer vision
  and pattern recognition}, 2016, pp. 770--778.

\bibitem{worrall2017interpretable}
D.~E. Worrall, S.~J. Garbin, D.~Turmukhambetov, and G.~J. Brostow,
  ``Interpretable transformations with encoder-decoder networks,'' in
  \emph{Proceedings of the IEEE International Conference on Computer Vision},
  2017, pp. 5726--5735.

\bibitem{li2012robust}
S.~Li, C.~Xu, and M.~Xie, ``A robust o (n) solution to the perspective-n-point
  problem,'' \emph{IEEE transactions on pattern analysis and machine
  intelligence}, vol.~34, no.~7, pp. 1444--1450, 2012.

\bibitem{eslami2018neural}
S.~A. Eslami, D.~Jimenez~Rezende, F.~Besse, F.~Viola, A.~S. Morcos, M.~Garnelo,
  A.~Ruderman, A.~A. Rusu, I.~Danihelka, K.~Gregor \emph{et~al.}, ``Neural
  scene representation and rendering,'' \emph{Science}, vol. 360, no. 6394, pp.
  1204--1210, 2018.

\end{thebibliography}

\clearpage
\newpage
\section*{Supplementary Material}

\subsection{System Settings:} 
\label{SM System Settings}

\textbf{Important Parameters:} As discussed in Section \ref{Simulation Settings}, in each data collection or evaluation episode, the robot first moves the camera to the random sampled desired pose $^{o} \boldsymbol{T}_{c^{*}}$. After collecting the desired observation by the camera, the robot moves the camera to the random sampled initial pose $^{o} \boldsymbol{T}_{c}$. Then the robot will try to position the camera back to the desired pose $^{o} \boldsymbol{T}_{c^{*}}$ under the guidance of VS controller. In order to ensure safety, the amplitude of linear velocity $v_c$ and angular velocity $w_c$ will be limited within 0.15m/s. An episode is considered to be finished successfully if the total error between the current  and the desired keypoints is lower than a specified threshold $\delta_f$:
\begin{equation}
    \sum _{k=1} ^n \left| u_k - u_k^* \right| + \left| v_k - v_k^* \right| \leq \delta_f
\vspace{-0.1cm} \end{equation}
Before the keypoint error fully converges, there are several situations that trigger the early termination of the episode:
\begin{itemize}
\item Every episode has a maximum steps of $\delta_s$ with each step takes $0.1s$. An episode will finish if the keypoint error hasn't reached the threshold $\delta_f$ within $\delta_s$ steps.
\item The camera walks out the workspace.
\item Any 2D keypoint is out of the camera's Fov.
\end{itemize}

We use four criterias to analysis the servo performance: servo success rate (\textbf{SR}), servo efficiency (timesteps,\textbf{TS}), final rotation error (\textbf{RE}) and final translation error (\textbf{TE}). We calculate the transformation between the final camera pose and the desired pose $^{{c}^*}\boldsymbol{T}_{c}$ by:
\begin{equation}
\label{RE and TE}
    ^{{c}^*}\boldsymbol{T}_{c} = \ ^{{c}^*}\boldsymbol{T}_o \cdot \ ^o \boldsymbol{T}_{c} = \left[ \begin{array}{cc}
        ^{{c}^*}\boldsymbol{R}_{c} & ^{{c}^*}\boldsymbol{t}_{c} \\
        \boldsymbol{0} & 1
    \end{array} \right]
\vspace{-0.1cm} \end{equation}
in simulation, and by:
\begin{equation}
    ^{{c}^*}\boldsymbol{T}_{c} = \ ^{{c}^*}\boldsymbol{T}_b \cdot \ ^b \boldsymbol{T}_{c} = \left[ \begin{array}{cc}
        ^{{c}^*}\boldsymbol{R}_{c} & ^{{c}^*}\boldsymbol{t}_{c} \\
        \boldsymbol{0} & 1
    \end{array} \right]
\vspace{-0.1cm} \end{equation}
in real world.

Rotation Error : The relative rotation $^{{c}^*}\boldsymbol{R}_{c}$ is converted into an axis-angle representation $\theta \boldsymbol{u}$. The rotation accuracy of camera is considered satisfactory if the deflection angle between the final pose and the desired pose is less than threshold $\delta_r$.

Translation Error : The translation accuracy of camera is considered satisfactory if the displacement between the final position and the desired position is less than $\delta_t$.
Tab.~\ref{table0_SystemSetting} shows specific parameter values. 

\begin{table}[htbp]
\centering
\caption{The specific value of parameters used in simulation and or real world settings.}
\label{table0_SystemSetting}
\begin{tabular}{c|c|c}
\hline
\hline
                & \bf{Value}                 & \bf{Units}           \\ \hline
$v_c$,$w_c$     & [-0.15,0.15]            & m/s             \\ \hline
$\delta_f$(sim)       & 10                    & pixel           \\ \hline
$\delta_s$(sim)       & 600                   & step            \\ \hline
$\delta_r$(sim)       & $\pi$/36              & rad             \\ \hline
$\delta_t$(sim)       & 0.866                 & cm               \\ \hline
$\delta_f$(real)      & 10                    & pixel           \\ \hline
$\delta_s$(real)      & 200                   & step            \\ \hline
$\delta_r$(real)      & $\pi$/36              & rad             \\ \hline
$\delta_t$(real)      & 1.732                 & cm               \\ \hline
\hline
\end{tabular}
\vspace{-0.2cm}
\end{table}

\begin{table}[htbp]
\centering
\caption{The 3D models of the target objects used in this paper: the specific coordinates of the 3D feature points in objects' own frame. The ground truth 2D keypoints annotations are labeled on real world observations according to the 3D models. They are also used to project 2D keypoints when training the neural controller in simulation. }
\label{table1_3Dmodel}
\begin{tabular}{c|c|c|c|c}
\hline
\hline
              & \bf{Feature Index}   & \bf{x(m)}       & \bf{y(m)}       & \bf{z(m)}      \\ \hline
Apriltag      & 1               & -0.0300 & -0.0300 & 0      \\ \hline
              & 2               & 0.0300  & -0.0300 & 0      \\ \cline{2-5}
              & 3               & 0.0300  & 0.0300  & 0      \\ \cline{2-5}
              & 4               & -0.0300 & 0.0300  & 0      \\ \hline
Charging Port & 1               & 0       & 0       & 0      \\ \hline
              & 2               & 0.0160  & 0       & 0      \\ \cline{2-5}
              & 3               & 0.0325  & 0       & 0      \\ \cline{2-5}
              & 4               & 0.0750  & -0.0150 & 0      \\ \cline{2-5}
              & 5               & 0.0240  & -0.0150 & 0      \\ \hline
Toy Horse     & 1               & 0.0320  & 0.0030  & 0.0250 \\ \hline
              & 2               & 0.0900  & 0.0030  & 0.0250 \\ \cline{2-5}
              & 3               & 0.1140  & 0.0850  & 0.0350 \\ \cline{2-5}
              & 4               & 0.0410  & 0.0680  & 0.0230 \\ \hline
\hline
\end{tabular}
\end{table}

\begin{figure}[htbp]
\centering
\includegraphics[width=\linewidth]{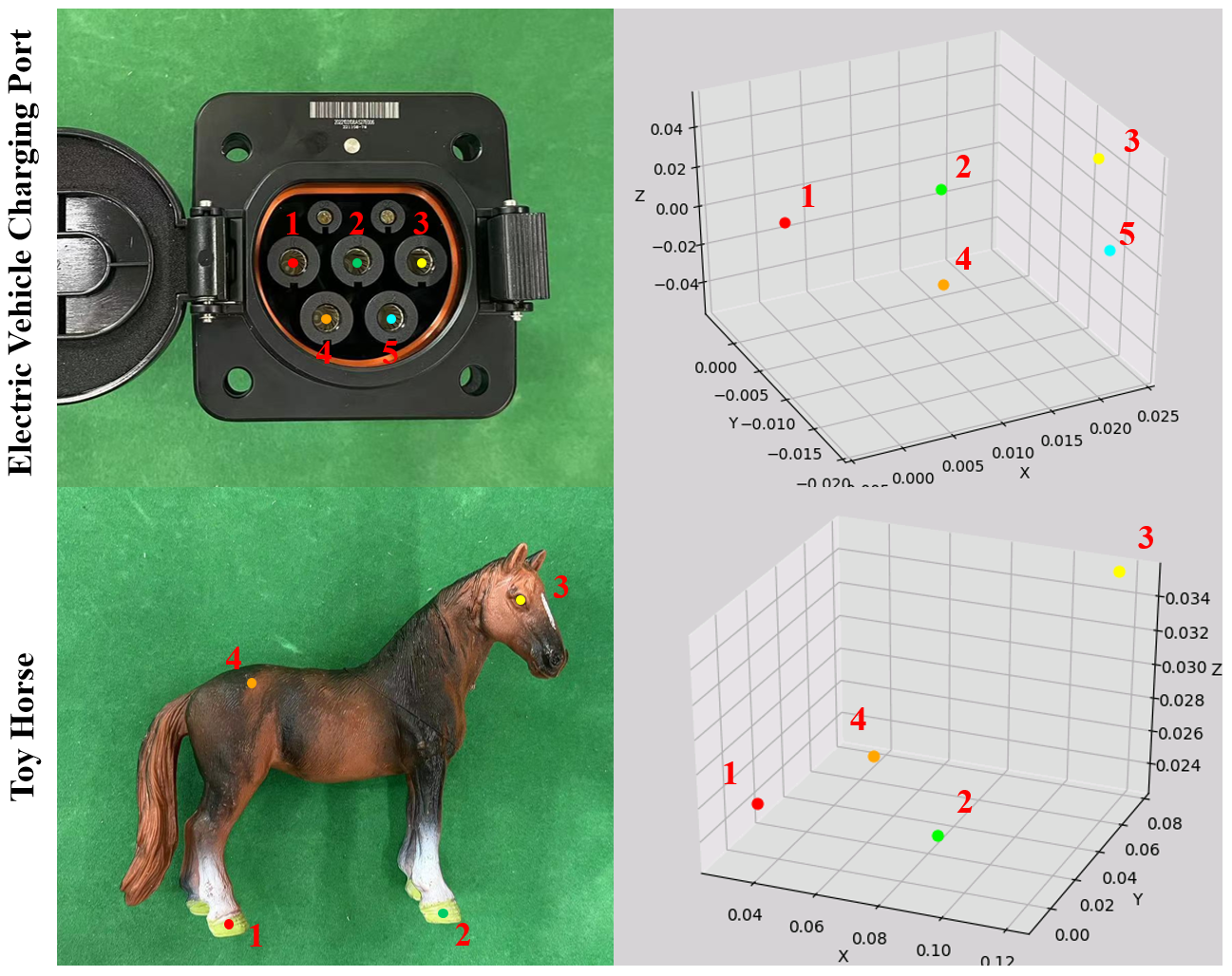}
\caption{Compared with some pioneer works that need accurate objects' meshes for rendering and training, we only need to define $n$ 3D feature points on the object and roughly measure their coordinates under object's frame as 3D model. This figure gives the visualization of 3D models of the target object used in real world: Five 3D feature points are defined at the center of round holes of the electric vehicle charging port. Four 3D feature points are defined at the toy horse's eye, hoof and back. }
\label{fig:fig14_NO_3Dmodel}
\vspace{-0.2cm}
\end{figure}

\begin{figure*}[htbp]
\centering
\includegraphics[width=0.9\linewidth]{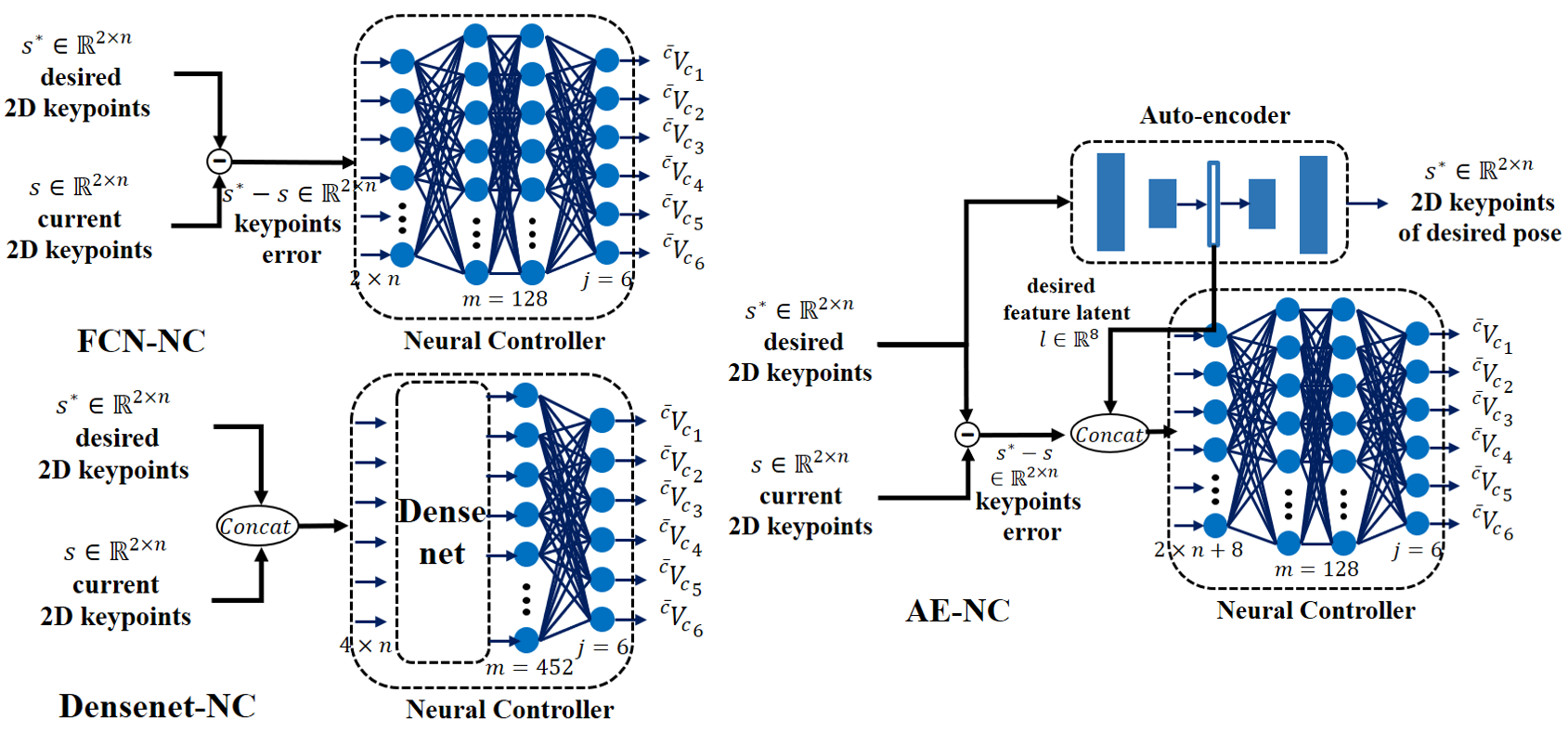}
\caption{The specific implementation of FCN-NC\cite{levine2016end}, DenseNet-NC\cite{puang2020kovis}, and AE-NC\cite{worrall2017interpretable}. These NCs have different structures but are all supervised by the same PBVS teacher.}
\label{fig:fig12_NC}
\vspace{-0.1cm}
\end{figure*}

\textbf{Target Objects' 3D Models:} Target objects' 3D models are used to project 2D keypoints on camera's plane to train NCs in simulation. To train the neural observer, the ground truth 2D keypoints annotations are labeled on real world observations according to the 3D models. According to \cite{chaumette2006basic}, users should define at least three 3D feature points on the real world objects for 6 DOF control and measure their positional relationship to get the 3D model. Usually, more than three 3D feature points are defined to avoid global minima. Fig.~\ref{fig:fig11_3Dmodel} shows several objects we used as the target objects in this work.  We define four 3D feature points on the apriltag's corners. The corners of apriltag are easy to identify and it is only used to compare the performance of different controllers in simulation. Real world experiments are performed on objects without obvious feature: a electric vehicle charging port and a toy horse. Five 3D feature points of electric vehicle charging port are defined at the center of round holes. Four 3D feature points of toy horse are defined at the horse's eye, hoof and back. Tab. \ref{table1_3Dmodel} shows the specific coordinates of the 3D feature points.



\subsection{Neural Controller Architecture:} 
\textbf{FCN-NC:} Fully connected neural controller (FCN-NC) is derived from the neural controller of \cite{levine2016end}. As shown in Fig.~\ref{fig:fig12_NC}, it is a three layers fully connected neural network with the relu activation for the first two layers and the tanh activation for the last layer, which shares the same architecture with the lower neural controller of HPN-NC. The input and output of FCN-NC are the same as the traditional IBVS controller. Its input is the 2D keypoints error between desired 2D keypoints and current 2D keypoints, and the output is a 6-dimensional vector containing linear velocity and angular velocity. FCN-NC is supervised by PBVS controller so its control performance is better than that of IBVS controller: As shown in Tab. \ref{table2_simulation_nc}, FCN-NC's SR is nearly twice than that of IBVS, and RE and TE are also smaller than IBVS. 27.8$\%$ SR is still low for practical use, indicating that FCN-NC is unable to adapt to multiple desired poses. This is caused by two reasons: One is the lack of desired pose information with 2D keypoints error input. The other is that the architecture of FCN-NC is too simple and the learning ability is limited. 

\textbf{DenseNet-NC:} DenseNet based neural controller (DenseNet-NC) is the controller used in \cite{puang2020kovis}. Its main structure is a DenseNet following with a fully connected layer. Compared to FCN-NC, it is improved in two ways. Firstly, DenseNet-NC has a more complex architecture than fully connected neural networks, which leads to stronger learning ability. Secondly, taking the 2D keypoints error as input like IBVS and FCN-NC lacks of desired pose information. To add the information of the desired pose, we modify the input of DenseNet-NC to be the concatenated vector of desired pose and current pose's 2D keypoints, as shown in Fig.~\ref{fig:fig12_NC}. Therefore, compared with FCN-NC, SR of DenseNet-NC is improved to 87$\%$, and the servo accuracy is also higher. However, due to the complex network structure and enlarged input, the training time and the duration of single forward prediction is longer than other networks (shown in Tab. \ref{table2_simulation_nc}). Simply adding the information of the desired pose to input and making the network structure more complex is not the optimal solution. 

\textbf{AE-NC:} Auto-encoder based neural controller (AE-NC) draws on the structure of auto-encoder\cite{worrall2017interpretable} to servo arbitrary desired pose. As shown in Fig.~\ref{fig:fig12_NC},  except for a low-level controller similar to FCN-NC, an auto-encoder\cite{worrall2017interpretable,eslami2018neural} is dedicated to encode the information of the desired pose. The input of the auto-encoder is desired 2D keypoints and the auto-encoder reconstructs the input with a four layers fully connected neural networks. The first two layers output an 8-dimensional latent vector as encoded information of desired poses, and the last two layers of the network will restore the input desired 2D keypoints with the latent vector. The remaining structure of AE-NC is similar to FCN-NC: a three layers fully connected neural network is responsible for control inference, with a $2n+8$ dimensional vector concatenated by the 8-dimensional latent vector and the 2D keypoint error as input. As shown in Tab.~\ref{table2_simulation_nc}, AE-NC has higher SR and servo accuracy than DenseNet-NC with a relatively lightweight neural controller. This result shows that the servo process itself is not difficult to learn but the effectiveness of modeling the desired poses' information has a significant impact on servo performance. Thus enhancing the controller network is not necessary but a powerful desired pose modeling mechanism is very important. Regrettably, compared with FCN-NC and DenseNet-NC, AE-NC needs to train an additional auto-encoder network and freeze its parameters before training the low-level neural controller, which undoubtedly increases the workload of training. Moreover, the input of AE-NC has been enlarged by the 8-dimensional latent vector, which also increases the number of neural controller's parameters and prolongs the inference time. This is why we choose HPN-NC as the final solution. HPN-NC can be trained together as a whole. Because the upper HPN will not participate in the VS process after the inference of lower NC's parameters, it will not enlarge the parameter amount of the lower NC.

\subsection{Hyper-network Neural Controller:}

\begin{figure}[t]
\centering
\includegraphics[width=0.95\linewidth]{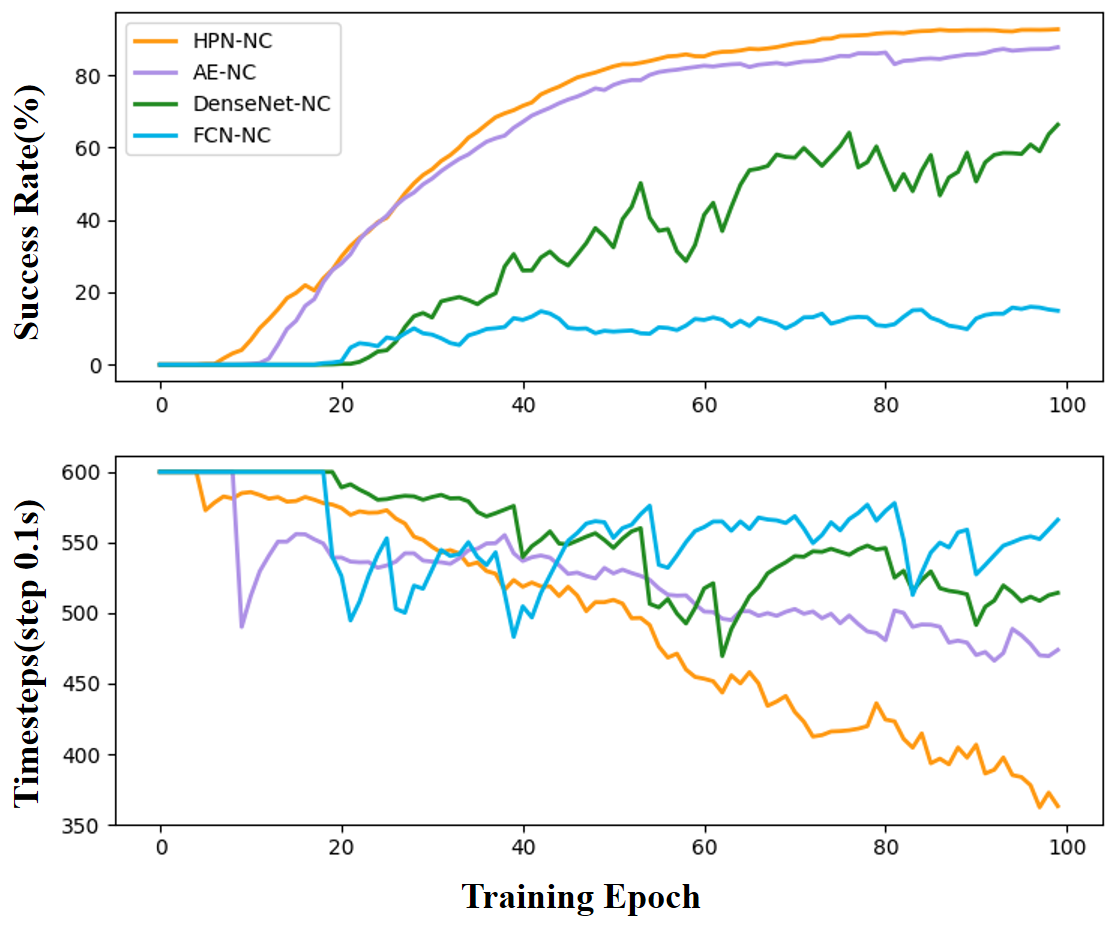}
\caption{The SR and servo efficiency(TS) curve of 4 NCs. As HPN-NC and AE-NC explicitly encode the desired poses, they have obvious advantages(higher SR and smaller TS) over FCN-NC and DenseNet-NC. HPN-NC outperforms AE-NC for a stronger modulate mechanism.}
\label{fig:fig6_NC}
\vspace{-0.35cm}
\end{figure}

\begin{figure}[t]
\centering
\includegraphics[width=0.9\linewidth]{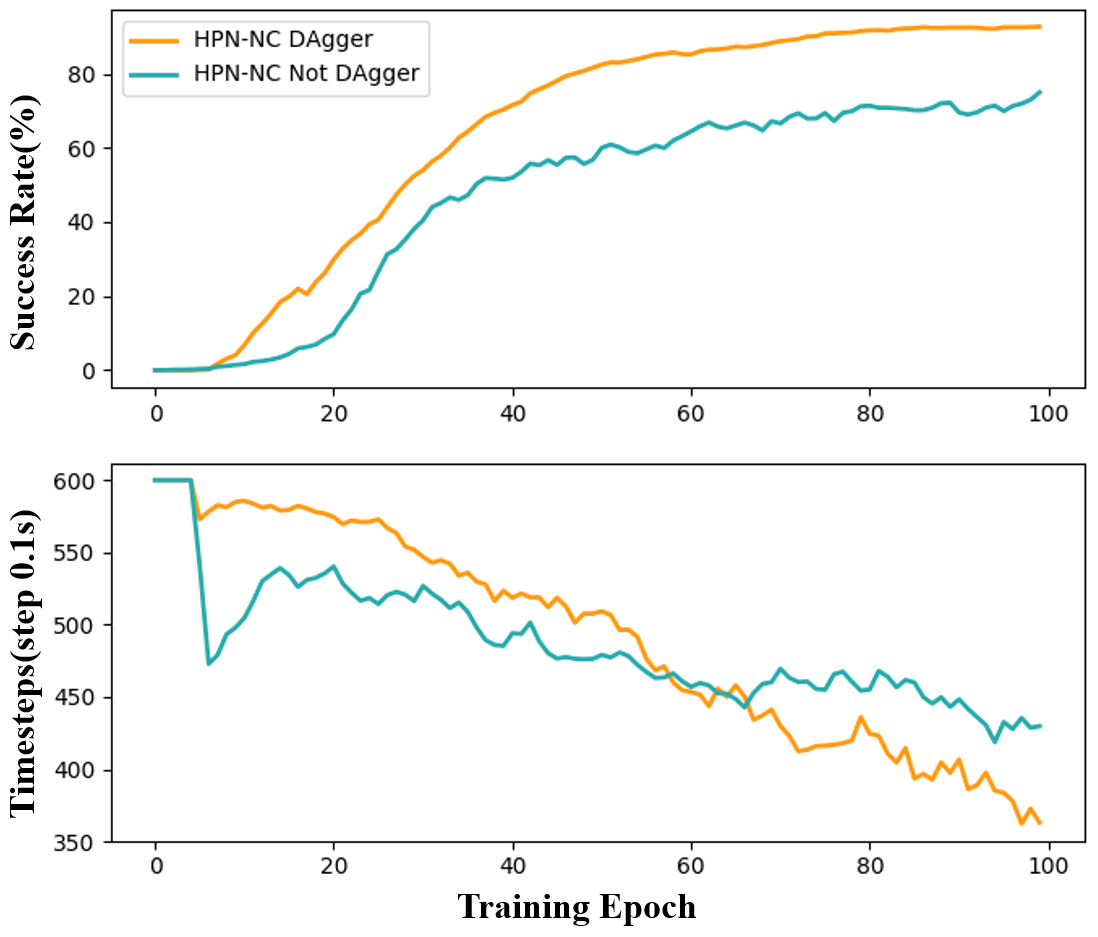}
\caption{The success rate and and servo efficiency(timesteps) curves of DAgger and without DAgger (imitation learning) for 100 epochs of training. DAgger helps the model to learn faster than imitation learning.}
\label{fig:fig4_DAgger}
\end{figure}

\begin{figure*}[htbp]
\centering
\subfigure[Task 1]{
\begin{minipage}[t]{0.501\linewidth}
\centering
\includegraphics[width=\textwidth]{fig/fig16_case1.png}
\label{Task 1}
\vspace{-0.4cm}
\end{minipage}%
}
\subfigure[Task 2]{
\begin{minipage}[t]{0.478\linewidth}
\centering
\includegraphics[width=\textwidth]{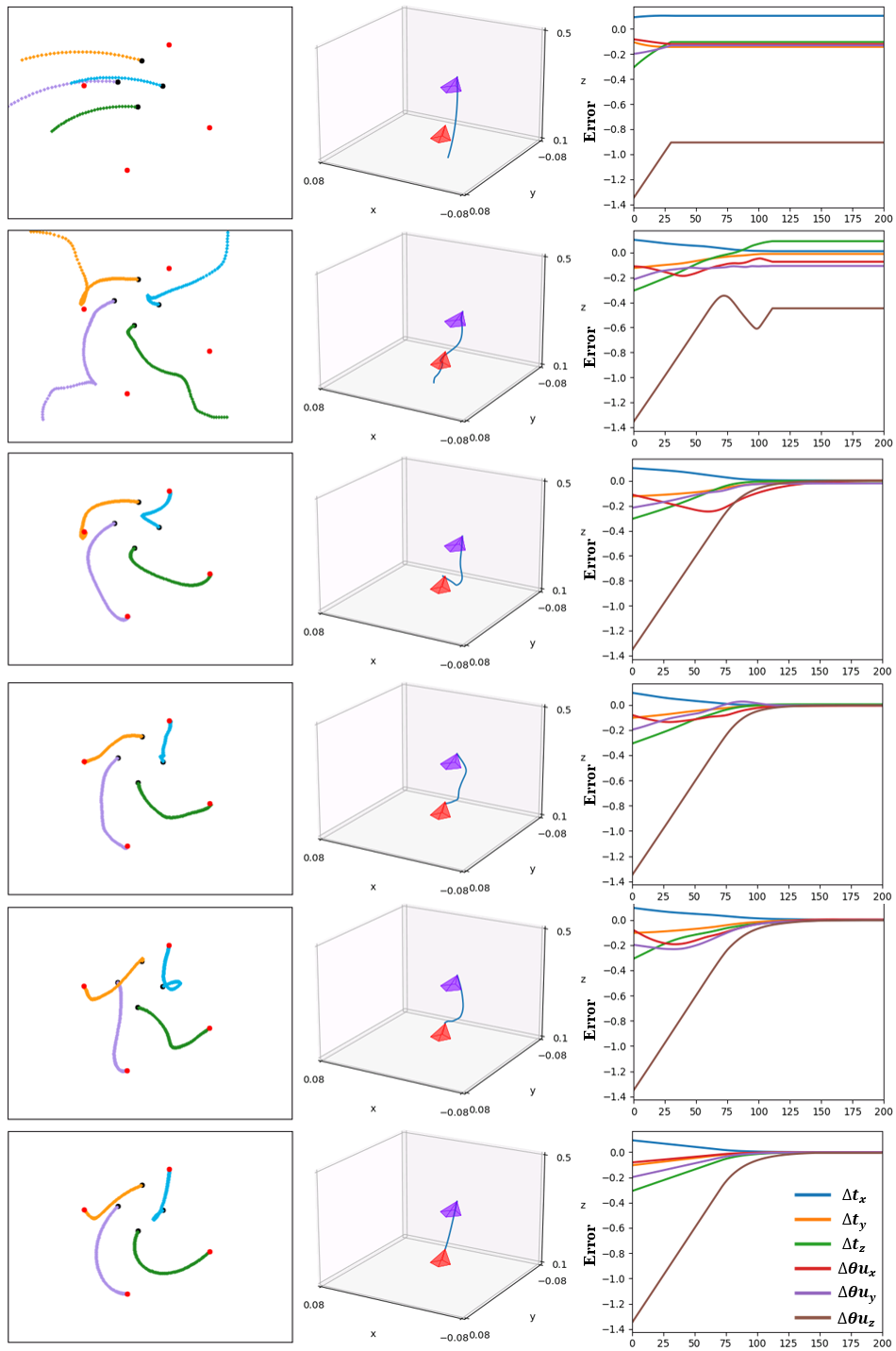}
\label{Task 2}
\vspace{-0.4cm}
\end{minipage}%
}
\centering
\caption{2D, 3D and error trajectories for two VS tasks. 
For 2D trajectories, black dots represent the initial keypoints and red dots represent the desired keypoints. For 3D trajectories, purple triangles represent the initial camera poses and red triangles represent the desired camera poses. Error trajectories visualize the TE and RE between the current and the desired camera poses. TE is in meters and RE is represented by the axis angle.}
\label{fig:fig16_case12}
\vspace{-0.15cm}
\end{figure*}

\textbf{Comparison with Other Controller:} Fig.~\ref{fig:fig16_case12} shows the visualization of 2D, 3D and error trajectories for two visual servoing tasks. These two VS tasks have different desired poses and initial poses. For 2D trajectories, black dots represent the initial keypoints and red dots represent the desired keypoints. For 3D trajectories, purple triangles represent the initial camera poses and red triangles represent the desired camera poses. Error trajectories visualize the translation and rotation error between current camera poses and desired camera poses, which are calculated by Eq.~\ref{RE and TE}. The rotation error is represented by the axis angle. The controllers are IBVS\cite{chaumette2006basic}, FCN-NC\cite{levine2016end}, DenseNet-NC\cite{puang2020kovis}, AE-NC\cite{worrall2017interpretable}, HPN-NC and ground truth PBVS\cite{chaumette2006basic}. Due to the non-linearity of IBVS, its 2D and 3D trajectories have relatively large curvature. The curved trajectory causes the features to easily move out of the camera Fov. Especially when the number of feature correspondences is small(four or five for example), IBVS is easy to fail, leading to only 17.2$\%$ SR of IBVS. NCs learn how to act from PBVS, whose servo trajectory in 3D space is a straight line, so they act more secure in 3D space like PBVS. Task 1 is relatively simple: the initial TE does not exceed 0.3m and RE does not exceed 0.3. All controllers have reached the desired pose. But only DenseNet-NC, AE-NC and HPN-NC complete within 200 steps. The terminal error of HPN-NC is smaller than that of DenseNet-NC and AE-NC. The second task is more difficult, the initial $\mathbf{\theta}\mathbf{u}_z$ is -1.3 which means that there exists a large offset between the initial and the desired pose. IBVS and FCN-NC cannot complete VS at this time. DenseNet-NC and AE-NC's terminal error cannot be completely eliminated. HPN-NC still successfully completes the task.  

Fig.~\ref{fig:fig6_NC} shows the SR and servo efficiency curve TS of four models in training. FCN-NC is a simple fully connected neural networks and lacks information of the desired pose, so its performance is the worst. DenseNet-NC uses the DenseNet as the main structure which is a more complex architecture than FCN-NC and leads to stronger learning ability. Moreover, it has an independent input for desired keypoints. Therefore, its servo SR is greatly raised and the servo accuracy is also higher. However, due to the complex structure and enlarged input, the training time and the duration of single forward prediction is longer than other NCs. HPN-NC and AE-NC have obvious advantages over FCN-NC and DenseNet-NC. This advantage is derived from the encoding of desired poses' information with independent networks, which helps the controller better adjust to different desired poses. AE-NC\cite{worrall2017interpretable} uses an additional auto-encoder to get a latent vector that containing desired poses' information. HPN-NC no longer needs to train an encoding network separately. It uses a more complex hyper network to model the modulate mechanism of the VS task which can be trained end-to-end with the low-level neural controller. HPN-NC generates a lightweight low-level fully connected NC identical to FCN-NC given the desired keypoints before the servo process, which also ensures efficient inference. Compared with AE-NC, HPN-NC has even higher SR, servo accuracy (RE and TE) and shorter inference time. With such strong hyper net, HPN-NC's 2D and 3D trajectories are more similar to the ground truth PBVS's. 

\textbf{DAgger or Not:} We examine the effect of DAgger on HPN-NC training. Fig.~\ref{fig:fig4_DAgger} shows that DAgger enables the controller to learn faster and better than imitation learning. The model trained by DAgger has higher SR and servo accuracy after 100 epochs. VS is a sequential decision-making process, where a bad action at time t can lead to an unseen state. DAgger helps to address this out-of-distribution problem. It continuously expands the training dataset by interacting with the simulated environment until the controller’s performance stops improving.

\begin{figure*}[htbp]
\centering
\includegraphics[width=0.9\linewidth]{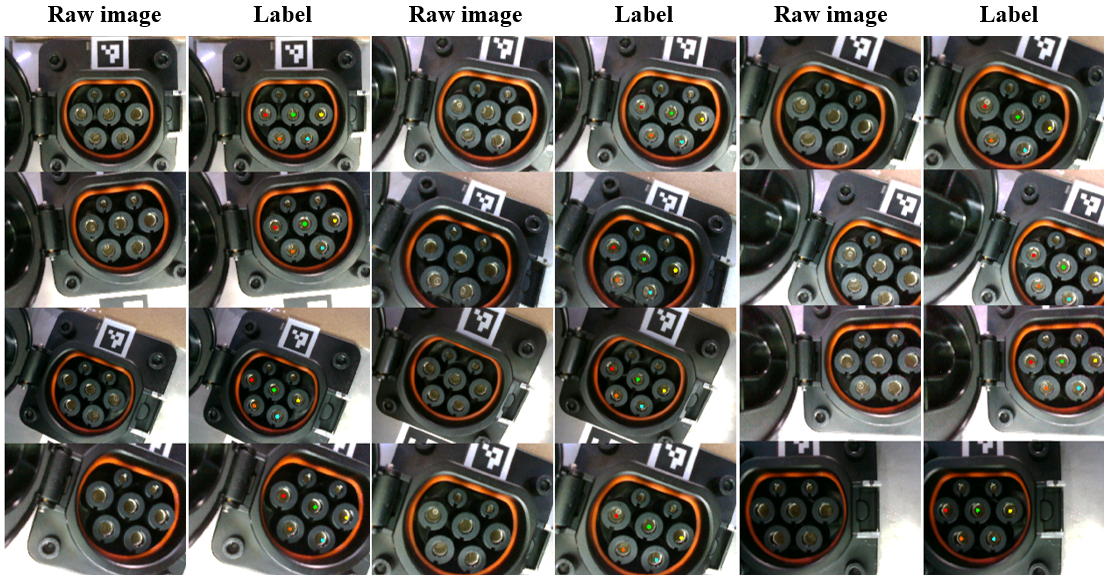}
\caption{Manual annotations we used to train NO of the charging port.}
\label{fig:fig13_NO_traindata}
\end{figure*}

\begin{figure*}[htbp]
\centering
\includegraphics[width=0.9\linewidth]{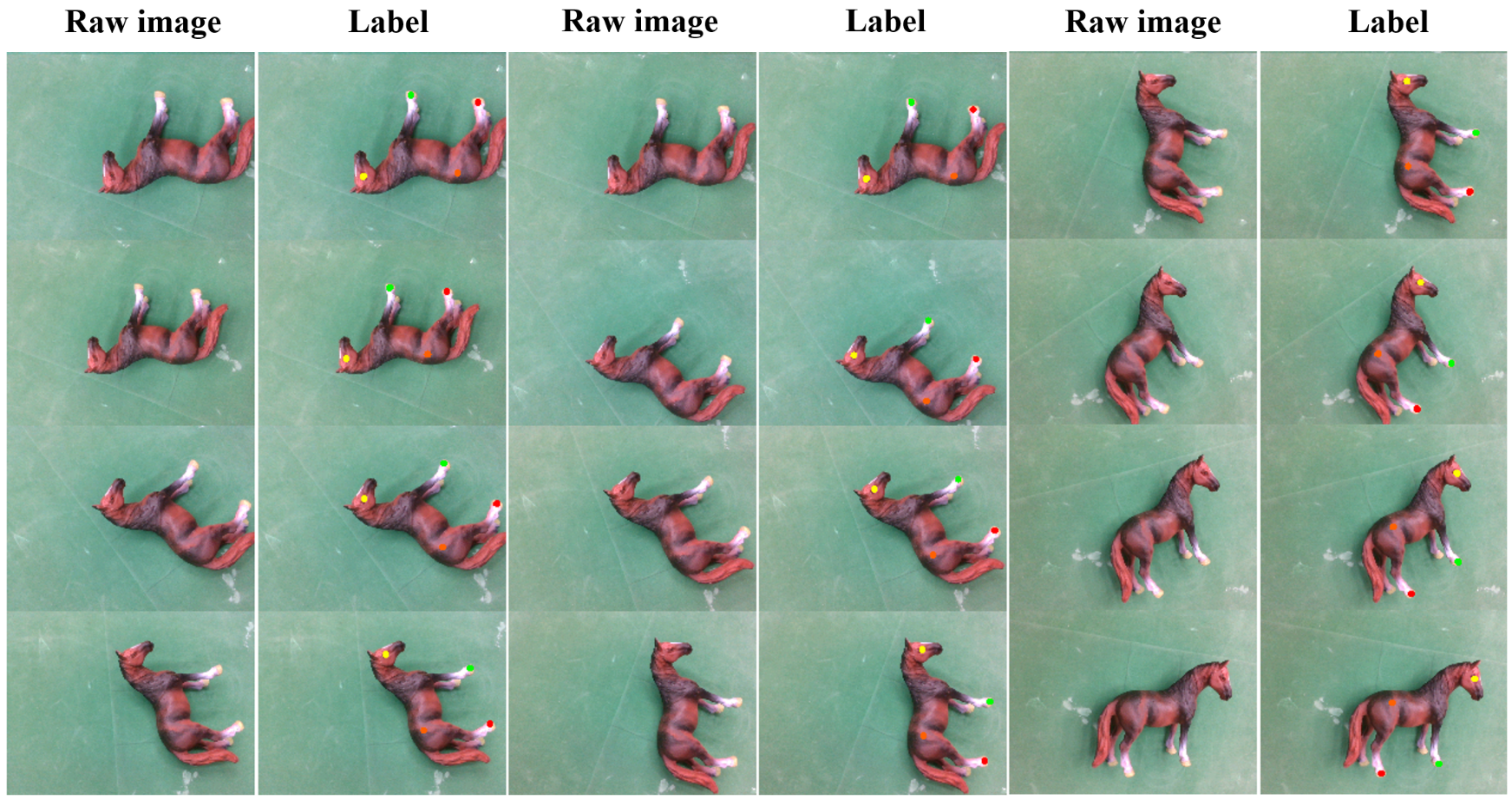}
\caption{Manual annotations we used to train NO of the toy horse.}
\label{fig:fig13_NO_traindata2}
\end{figure*}

\begin{table*}[h]
\centering
\caption{Performance comparison of HPN-NC and DenseNet-NC with the unsupervised observer proposed by \cite{puang2020kovis}. The comparison is performed on the water cup used in the original paper\cite{puang2020kovis} for 500 evaluations in simulation.}
\label{table3_Unsupervised}
\begin{tabular}{c|c|c|c|c}
\hline
\hline
\textbf{Controller}     & \textbf{SR(\%)} & \textbf{TS(0.1s)} & \textbf{RE(rad)} & \textbf{TE(cm)} \\ \hline
DenseNet-NC\cite{puang2020kovis}                   & 56                        & 213.44                   & 0.096           & \bf{1.136}\\ \hline
\bf{HPN-NC}         & \bf{81.4}                 & \bf{206.22}              & \bf{0.066}     & 1.586     \\ \hline \hline
\end{tabular}
\end{table*}

\subsection{Hyper-network Neural Controller with Unsupervised Neural Observer:} 

In order to solve the problem of high cost of manual annotating 2D keypoints, \cite{puang2020kovis} provides an unsupervised  neural observer for visual servoing (VS). Given the mesh of target object, it learns the neural observer (NO) and neural controller (NC) in an self-supervised manner in simulation and can directly transfer the model to the real world to complete robotic manipulation tasks. We modify the VS task to have multiple desired poses. We use HPN-NC to servo the water cup (one of the target objects used in the original manuscript) and compare HPN-NC with DenseNet-NC used in original paper \cite{puang2020kovis} given the same unsupervised neural observer. The experimental results show that HPN-NC can be used in conjunction with the unsupervised NO proposed by \cite{puang2020kovis} to servo arbitrary desired poses. As shown in Tab. \ref{table3_Unsupervised}, HPN-NC achieves 81.4$\%$ servo SR and has a higher servo efficiency compared with the original DenseNet-NC. 

The reported servo SR of DenseNet-NC in \cite{puang2020kovis} is 100$\%$ to servo fix desired poses. It drops to 56$\%$ for the multiple desired poses visual servoing task. This degradation is mainly caused by inconsistent 2D keypoints. Keypoints imply the information of relative pose between the object and the camera, while the consistency of keypoints between different viewing perspectives is a necessary guarantee for successful VS. The unsupervised keypoint extractors proposed by \cite{puang2020kovis} is conditioned on 4 DOF VS task with fix desired pose and small initial pose offset: the maximum initial pose offset between the initial and desired pose is $\Delta \mathbf{r}_{0}=(^{{c}^*}\mathbf{t}_{c},\mathbf{\theta}\mathbf{u}):$ $^{{c}^*}\mathbf{t}_{c} = (1.5cm,1.5cm,1.5cm), \mathbf{\theta}\mathbf{u} = (0\degree,0\degree,5\degree,)$. While in this letter, a random desired camera pose is sampled in the space 15cm above the target object with 0 to 5cm disturbance in $XYZ$ translation and the maximum initial pose offset between the initial and desired pose is $\Delta \mathbf{r}_{0}=(^{{c}^*}\mathbf{t}_{c},\mathbf{\theta}\mathbf{u}):$ $^{{c}^*}\mathbf{t}_{c} = (15cm,15cm,30cm), \mathbf{\theta}\mathbf{u} = (53.1\degree,53.1\degree,180\degree,)$. 

Keypoints extracted in an unsupervised manner cannot be geometrically consistent with such large offset. As a result, to servo desired poses in 6 DOF with large initial offset, the controller is enforced to encode all the correspondences between different desired poses and different keypoints, which may causes the performance degradation. In addition, the training of unsupervised observers requires the 3D mesh of target objects to render depth maps in simulation which is not always available in practical. As a result, to servo real world objects, although no manual cost is required to train an unsupervised keypoint extractor \cite{puang2020kovis}, we choose the supervised NO with human defined 2D keypoints as the keypoint extractor. 

\subsection{Hyper-network Neural Controller with Supervised Neural Observer:} 

\begin{figure}[t]
\centering
\includegraphics[width=0.95\linewidth]{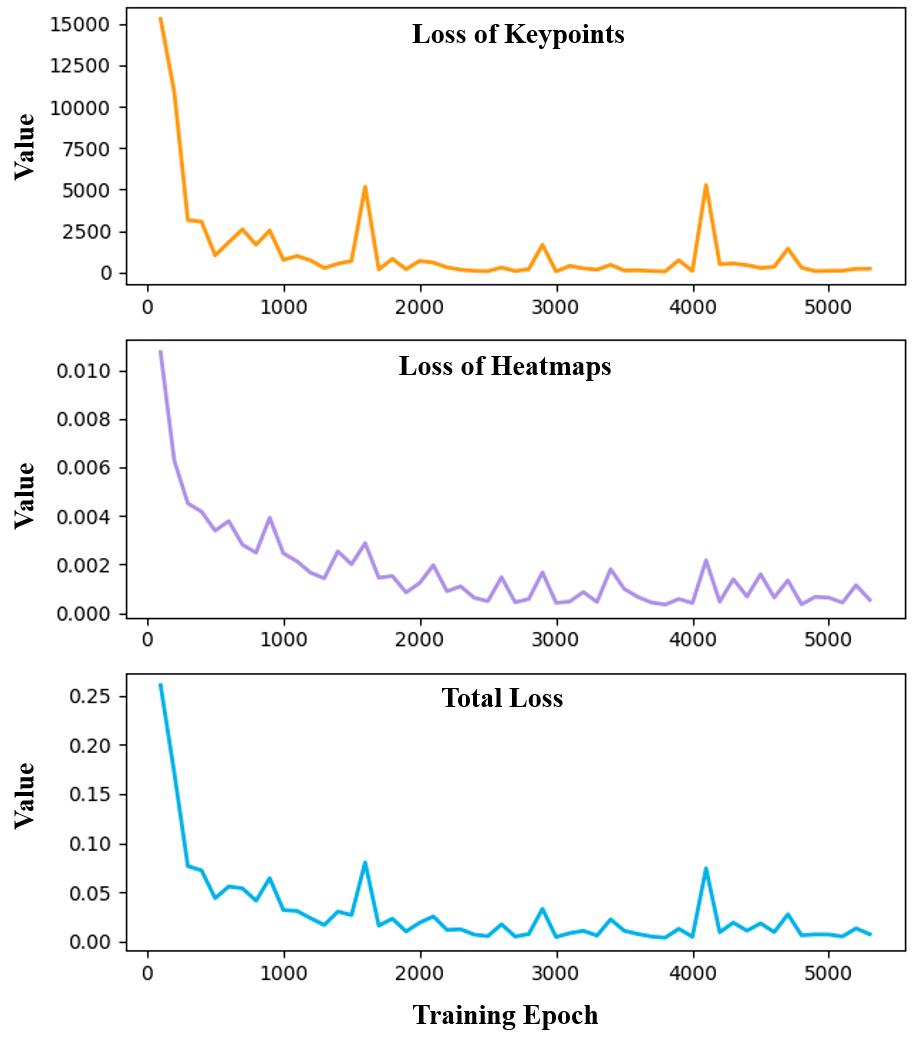}
\caption{The figure shows the loss curve of training a charging port's NO. The backbone is Resnet-18 and is trained on 600 annotations and evaluated on a test set composed of another 300 labeled data. The training loss consists of heatmaps' regression loss and keypoint coordinates' regression loss.}
\label{fig:fig15_NO_training_curve}
\end{figure}

\begin{figure}[t]
\centering
\includegraphics[width=0.999\linewidth]{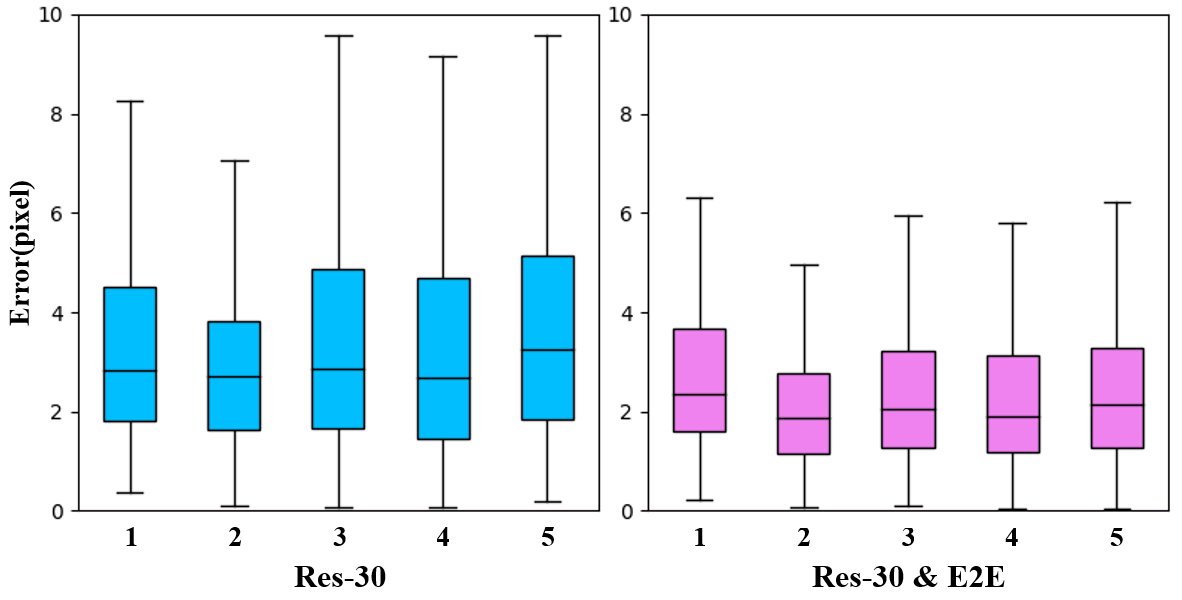}
\caption{The recognition error of each keypoint of Res-30 before and after self-supervised end-to-end training. The left part shows the recognition error of Res-30 trained with 30 mannual annotations. The right part shows the recognition error of Res-30 fine-tuned with end-to-end training. The average recognition error of each keypoint is reduced from three pixels to two pixels after training. }
\label{fig:fig15_NO_prediction_error4}
\end{figure}

\begin{figure}[t]
\centering
\includegraphics[width=\linewidth]{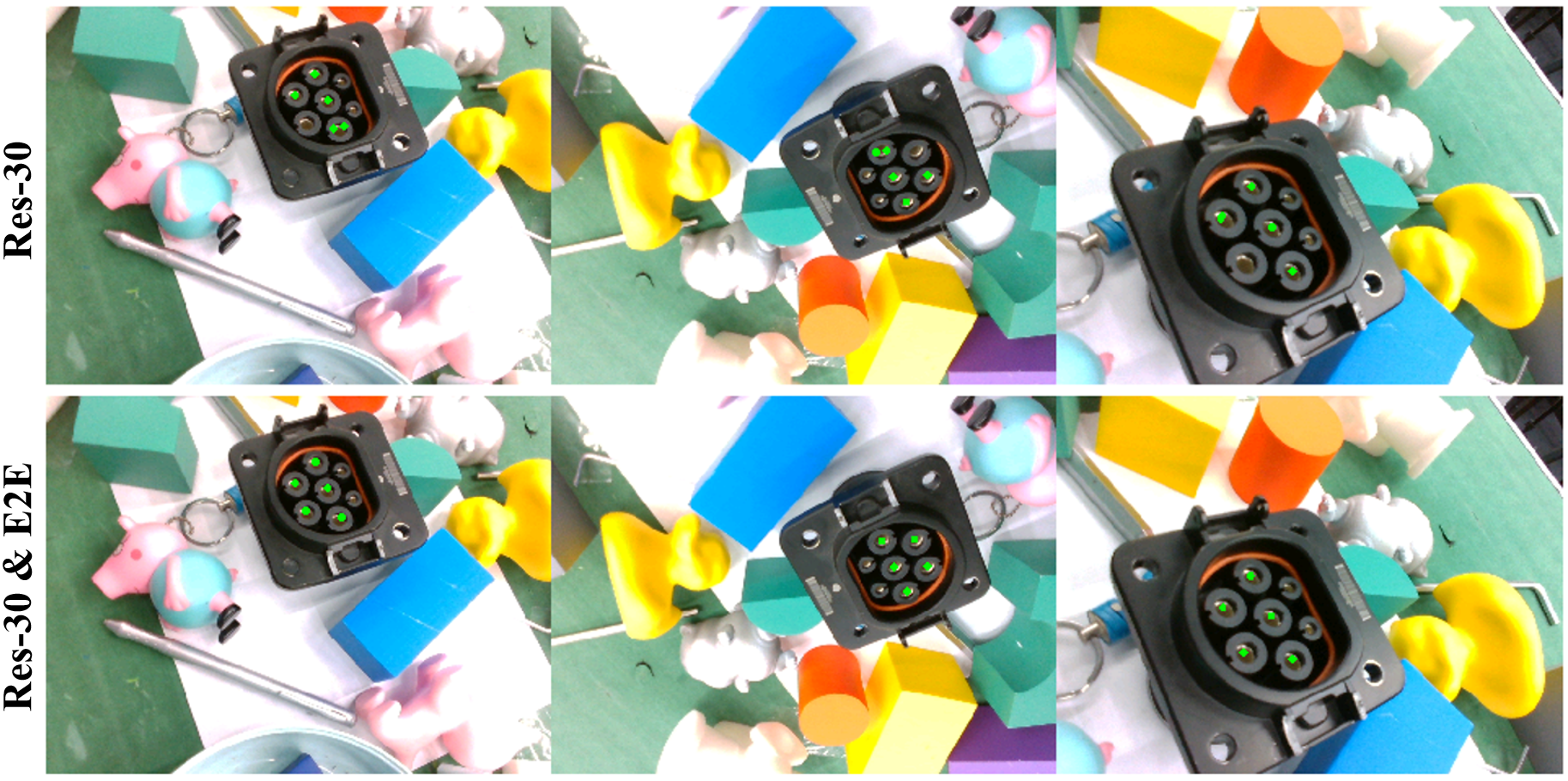}
\caption{Comparison of keypoint extraction ability before and after end-to-end training. The green pixels represent the 2D keypoint predicted by NO. The first row shows the performance of Res30. The extracted keypoints have obvious error in some perspective:some keypoints deviate from the center of the circular hole. After self-supervised end-to-end training, Res30E has correct prediction in the same perspective.}
\label{fig:fig9_e2e}
\end{figure}

\begin{table*}[t]
\centering
\caption{Performance comparison of integrated model with different mannual annotations. The neural observers are ResNet-18 trained with different amount of mannual annotations. The comparison is performed on a charging port shown in Fig.~\ref{fig:fig11_3Dmodel} for 50 evaluations in real world.}
\label{table5_annotations}
\begin{tabular}{l|c|c|c|c}
\hline
\hline
\textbf{Integrated Model}& \textbf{SR(\%)} & \textbf{TS(0.1s)} & \textbf{RE(rad)} & \textbf{TE(cm)} \\ \hline
Res-600 PBVS-PnP      & 88        & 90.29             & 0.142                & 2.950               \\ \hline
Res-300 PBVS-PnP      & 56        & 92.00             & 0.115                & 2.391               \\ \hline
Res-100 PBVS-PnP      & 42        & 93.19             & 0.153                & 3.179               \\ \hline
Res-60  PBVS-PnP      & 36        & 96.26             & 0.122                & 2.463               \\ \hline
Res-30  PBVS-PnP      & 30        & \bf{89.06}        & 0.266                & 4.365               \\ \hline
Res-30 HPN-NC $\&$ E2E    & \bf{92}                        & 112.48                   & \bf{0.059}         & \bf{1.275}          \\ \hline
\hline 
\end{tabular}
\vspace{-0.1cm}
\end{table*}

\begin{table*}[t]
\centering
\caption{Performance comparison of integrated model. The neural observers are ResNet-18 trained with 10 mannual annotations of the toy horse. The comparison is performed on the toy horse shown in Fig.~\ref{fig:fig11_3Dmodel} for 50 evaluations in real world.}
\label{table6_toyhorse}
\begin{tabular}{l|c|c|c|c}
\hline
\hline
\textbf{Integrated Model}& \textbf{SR(\%)} & \textbf{TS(0.1s)} & \textbf{RE(rad)} & \textbf{TE(cm)} \\ \hline
Res-10 HPN-NC      & 88        & 115.34             & 0.057                & 1.699               \\ \hline
\bf{Res-10 HPN-NC $\&$ E2E} & \bf{90}   & \bf{112.26}            & \bf{0.029}                & \bf{0.880}                      \\ \hline
\hline 
\end{tabular}
\vspace{-0.1cm}
\end{table*}

In this paragraph, we introduce how to train the supervised neural observer (NO) with manual annotations to servo real world objects. The main content includes training dataset preparation, training procedure, and performance improvement with self-supervised end-to-end training.

\textbf{Data Preparation:} For training data preparation, we annotate 2D keypoints on RGB images collected in real world to generate dataset $D_{NO}$ according to the pre-defined 3D model. By clicking on the image with the mouse, we mark the coordinates of the 2D keypoint. Manually annotated 2D keypoints simplify the encoding of correspondence between desired poses and keypoints. For example, for the charging port, we collect a dataset with 1000 manual annotations. During training neural observer with $n$ manual annotations, we will randomly sample $n$ manual annotations
from 1000 manual annotations and divide 80$\%$ of $n$ manual annotations as the training set and 20$\%$ of $n$ manual annotations as the validation set. Fig.~\ref{fig:fig13_NO_traindata} shows some of the manual annotations we used to train the NO of the charging port. These RGB images have 640x480 pixels. 2D keypoints are marked in different colors according to the pre-defined feature points of the 3D model given in Section \ref{SM System Settings}. Data augmentation operations including background replacement, translation, scaling, rotation, and homography matrix stretching are applied to this annotations during training to improve the generalization performance of NO.  

\textbf{Training Procedure:} The model will be trained for 5000 epochs and be evaluated every 100 epochs. Each epoch randomly samples 5 batch of images from the dataset with a batch size of 8. Fig.~\ref{fig:fig15_NO_training_curve} shows the training curve of a charging port's NO whose backbone is Resnet-18 and is trained on 600 manual annotations (the model in the first line of Tab.~\ref{table4_MAR}). As discussed in Section \ref{threestage}, the training procedure of NO has two goals. Firstly, NO tries to minimize the difference between the predicted heatmap and the ground truth heatmap $g_i(x)$ peaking at $\boldsymbol{s}^{MA}$. At the same time, in order to improve the accuracy of predicted keypoints, NO tries to minimize the $\boldsymbol{L_2}$ norm between the keypoint coordinates calculated by spatial-softmax operation and the ground truth coordinates $\boldsymbol{s}^{MA}_{i}$. So the training loss consists of heatmaps' regression loss and keypoint coordinates' regression loss. Scale factors $\gamma_{\mathit{h} }=10$ and $\gamma_{\mathit{k} }=0.00001$ balance two loss. Eq. \ref{equ:loss_NO} gives the specific composition of the training loss. The NO with the smallest total loss will be selected as the real world keypoint extractor.

 

\begin{figure*}[htbp]
\centering
\includegraphics[width=\linewidth]{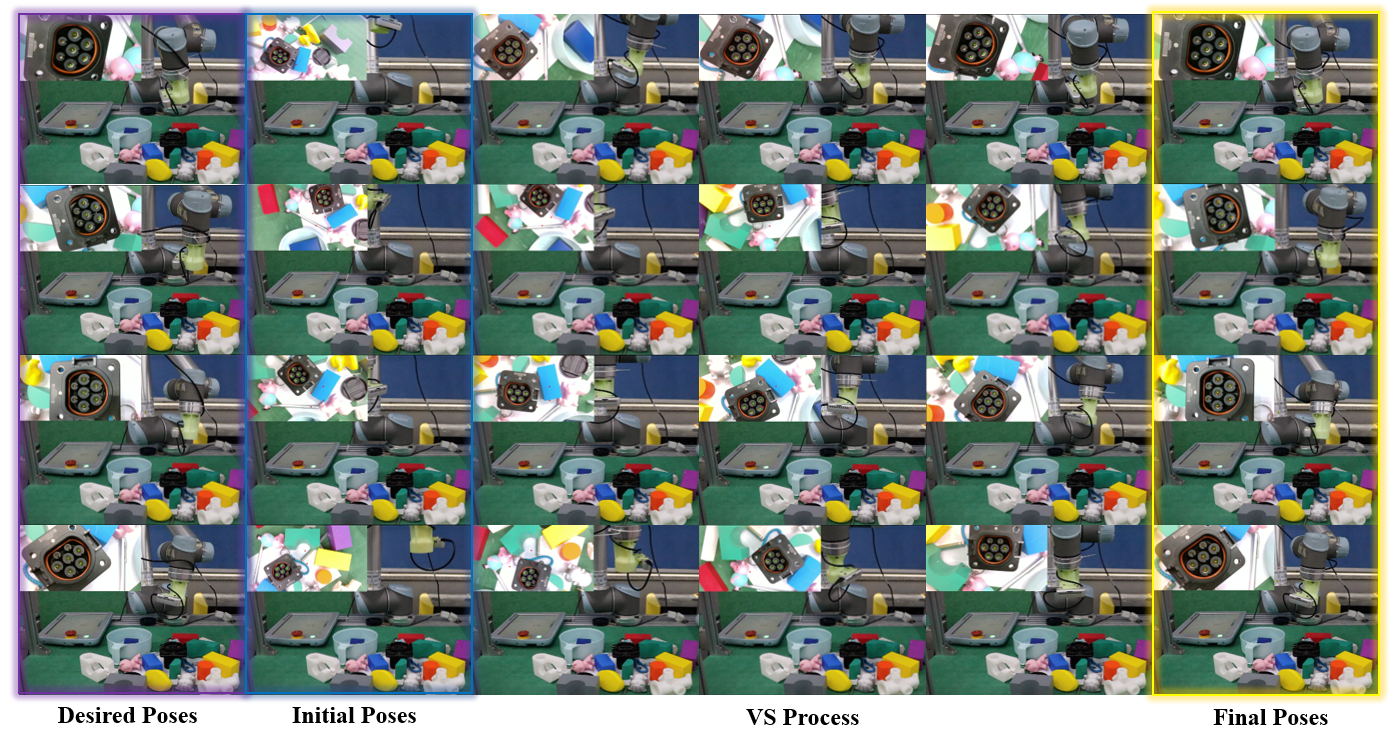}
\caption{Servo arbitrary desired poses with HPN-NC in real world, even those with large initial pose offset (e.g. 180$\degree$ rotation in yaw). The small window shows the observations of the camera. NO will extract 2D keypoints from these observations as input to HPN-NC. It can be seen that the background of the charging port is very different from the labels shown in Fig. \ref{fig:fig13_NO_traindata}, and NO can well adapt to the change of the background. More demos can be found in the attached video.}
\label{fig:fig9_3}
\end{figure*}

\textbf{Performance Improvement:} We obtain another 300 manual annotations to be the testing set for quantitative evaluation of different neural observers. Fig.~\ref{fig:fig15_NO_prediction_error3} shows the total recognition error of five NOs for the charging port on the testing set. The models from left to right are: NO trained with 600 annotations (Res-600), 300 annotations (Res-300), 30 annotations (Res-30), NO trained with 300 annotations and improved with self-supervised end-to-end data (Res-300 $\&$ E2E), NO trained with 30 data model and improved with self-supervised end-to-end data (Res-30 $\&$ E2E). These NOs are all evaluated on the testing set. The charging port's 3D model has five 3D feature points. For each feature point, we calculate the euclidean distance between predicted 2D keypoint coordinate and ground truth. All of the five distances are summed together as the total recognition error. Res-600 has smaller total recognition error than Res-300 and Res-30. A very intuitive idea is that as the training data decreases, the recognition error of NO will become larger and larger. After self-supervised end-to-end training, the performance of Res-300 and Res-30 can be improved. The total recognition error of both Res-300 $\&$ E2E and Res-30 $\&$ E2E reach the same level as Res-600. 

Fig.~\ref{fig:fig15_NO_prediction_error4} shows the recognition error of each keypoint of Res-30 before and after self-supervised end-to-end training. The average recognition error of each keypoint is reduced from three pixels to two pixels after training. Fig.~\ref{fig:fig9_e2e} gives some visualizations of keypoint recognition. Given the observation of the charging port from the same perspective, Res-30 $\&$ E2E has obviously better recognition performance.

Recognition error will affect the performance of VS. From Tab.~\ref{table5_annotations} we can see the impact of recognition error on the SR of VS. The neural observers are: NO trained with 600 annotations (Res-600), 300 annotations (Res-300), 100 annotations (Res-100), 60 annotations (Res-60), 30 annotations (Res-30). We connect these NOs trained by different numbers of manual annotations with the same controller, traditional PBVS controller, to form an integrated model. The relative poses of PBVS are estimated by Perspective-n-Point\cite{li2012robust} (PnP) with the 2D keypoints extracted by NOs. As the number of mannual annotations increases, the performance of the integrated model gets better: servo SR rises, and servo efficiency index TS drops. The integrated model of Res-600 and PBVS reachs $88\%$ SR. Nevertheless, PnP is not differentiable and cannot be trained in an end-to-end manner. Self-supervised end-to-end training can reduce recognition error and imporve VS performance. With fully differentiable HPN-NC, Res-30 improved with self-supervised end-to-end training (Res-30 $\&$ E2E) can achieve a SR higher than that of Res-600 with only 30 manual annotations.

\textbf{VS Tasks with Large Offset:}
Since our NO trained on manual annotations can maintain the consistency of 2D keypoints extraction in a relatively large range after self-supervised end-to-end training, we can achieve a large range VS. As shown in Fig.~\ref{fig:fig9_3}, some of the tasks have large initial pose offsets: maximum initial pose offset between the initial and desired pose is $\Delta \mathbf{r}_{0}=(^{{c}^*}\mathbf{t}_{c},\mathbf{\theta}\mathbf{u}):$ $^{{c}^*}\mathbf{t}_{c} = (15cm,15cm,30cm), \mathbf{\theta}\mathbf{u} = (53.1\degree,53.1\degree,180\degree,)$. The proposed method can still successfully complete the servo.

\subsection{Servo Other Real World Objects} 

Our proposed visual servoing model can generalize to different desired poses, but cannot generalize across different objects. This is because both neural observer and HPN-NC are trained based on a given 3D model. To VS a new object requires retraining NO and HPN-NC based on its 3D model. In this paragraph, we verify that our method can be applied to other real-world object without obvious features. The steps to servo a new object are as follows:
\begin{itemize}
\item Define the 3D feature points on the target object as 3D model. 

\item Train HPN-NC according to the 3D model under the supervision of PBVS in simulation.

\item Manual annotate 2D keypoints according to the 3D model in real world. Train NO on these manual annotations.

\item Fint-tune the integrated model of NO and HPN-NC with self-supervised end-to-end training in real world.
\end{itemize}

We choose the toy horse shown in Fig.~\ref{fig:fig14_NO_3Dmodel} as the target object. The toy horse has no obvious geometric structure to facilitate keypoints extraction. We defined four 3D feature points respectively at the horse's eye, hoof and back as the 3D model. We annotate 2D keypoints according to the pre-defined 3D model and provide a dataset with 170 manual annotations as shown in Fig.~\ref{fig:fig13_NO_traindata2}. We train a neural observer with 10 manual annotations and train a HPN-NC according to the 3D model in simulation. Then we further improved the integrated model consists of neural observer and HPN-NC with self-supervised end-to-end training described in Section~\ref{E2E}. Tab.~\ref{table6_toyhorse} gives the performance of the integrated model before and after self-supervised end-to-end training. It can be seen that although the success rate has not increased significantly, the servo accuracy(RE and TE) has improved. The success rate cannot be further improved because the horse model is larger than the charging port, so the feature points are easier to move out of the camera's FOV under the same settings. In contrast, the charging port has five similar circular hollows, which are more likely to cause misidentification. Therefore, when the neural observer are trained by few manually annotations, the servo success rate of the charging port is very low, and the improvement of the overall servo performance by end-to-end training is more obvious. 

\end{document}